\documentclass[a4paper,fleqn]{cas-dc} % Double Column

\usepackage[numbers]{natbib}
\usepackage[shortlabels]{enumitem}
\usepackage{hyperref}
\hypersetup{colorlinks = true,
    linkcolor=blue,
    filecolor=cyan,
    citecolor = blue,
    urlcolor=blue}%color hyperlinks
\usepackage[english]{babel}
\addto\extrasenglish{

} % Autoref Section not section%
\usepackage{algorithm} % algorithm package
\usepackage{algorithmic} % algorithm package
\usepackage{relsize} %Increase math font size in table
\usepackage{flushend} % Balance references
\usepackage{stfloats} % Table* at bottom

\def\tsc#1{\csdef{#1}{\textsc{\lowercase{#1}}\xspace}}
\tsc{WGM}
\tsc{QE}
\tsc{EP}
\tsc{PMS}
\tsc{BEC}
\tsc{DE}
\usepackage{setspace}
\begin{document}
\let\WriteBookmarks\relax
\def\floatpagepagefraction{1}
\def\textpagefraction{.001}

% %%%%%%%%%%%%%%%%%%%% TITLE PAGE %%%%%%%%%%%%%%%%%%%%%%%
% \begin{titlepage}
% \begin{center}
% \vspace*{10pt}
% \doublespacing
% \textbf{\Large CoTCoNet: An Optimized Coupled Transformer-Convolutional Network with an Adaptive Graph Reconstruction for Leukemia Detection}
% \vspace*{15pt}

% % Author names and affiliations
% Chandravardhan Singh Raghaw$^{a}$ (phd2201101016@iiti.ac.in), Arnav Sharma$^b$ (axs230011@utdallas.edu), \\ Shubhi Bansal$^a$ (phd2001201007@iiti.ac.in), Mohammad Zia
% Ur Rehman$^a$ (phd2101201005@iiti.ac.in), \\ Nagendra Kumar$^a$ (nagendra@iiti.ac.in) \\

% \hspace{10pt}
% \begin{flushleft}
% \small  
% $^a$ Indian Institute of Technology Indore, Indore 453552, India \\
% $^b$ The University of Texas at Dallas, Richardson, Texas, USA \\
% \vspace{1cm}
% \normalsize
% \textbf{Corresponding Author:} \\
% Chandravardhan Singh Raghaw \\
% Department of Computer Science and Engineering, \\
% Indian Institute of Technology Indore, Indore 453552, India \\
% Tel: +919887149322 \\
% Email: phd2201101016@iiti.ac.in

% \end{flushleft}        
% \end{center}
% \end{titlepage}

\begin{titlepage}
\doublespacing
{\centering{\Large \textbf{CoTCoNet: An Optimized Coupled Transformer-Convolutional Network with an Adaptive Graph Reconstruction for Leukemia Detection}}\\
Chandravardhan Singh Raghaw, Arnav Sharma, Shubhi Bansal, Mohammad Zia
Ur Rehman, Nagendra Kumar\\}

\vspace{2em}

{\large Highlights}

\begin{itemize}
\item An optimized novel coupled transformer-convolution network for leukemia detection.
\item Correlated global and spatial features help in identifying hematological malignancies.
\item Graph-based feature reconstruction module to capture hidden features of leukocytes.
\item A deep synthetic leukocyte generator is devised to mitigate data imbalance issues.
\item Extensive evaluations signify efficacy on four datasets with more than 16,982 WSIs.
\end{itemize}

\vspace{2em}

\noindent This is the preprint version of the accepted paper.\\
\noindent This paper is accepted in \textbf{Computers in Biology and Medicine, 2024.}
\\
DOI: \url{https://doi.org/10.1016/j.compbiomed.2024.108821}
\end{titlepage}
% %%%%%%%%%%%%%%%%%%%%%%%%%%%%%%%%%%%%%%%%%%%%%%%%%%%%%%%

% \shorttitle{CoTCoNet: An Optimized Coupled Transformer-Convolution Network for Leukemia Detection}
\shorttitle{CoTCoNet: An Optimized Coupled Transformer-Convolution Network for Leukemia Detection}

\shortauthors{Raghaw \textit{et~al.}}

\title [mode = title]{CoTCoNet: An Optimized Coupled Transformer-Convolution Network with an Adaptive Graph Reconstruction for Leukemia Detection}            
% \title [mode = title]{CoTCoNet: Leukemia Detection with an Optimized Coupled Transformer-Convolution Network with an Adaptive Graph Reconstruction leveraging GAN-simulated Leukocytes}                     

% \tnotemark[1,2]

% \tnotetext[1]{This document is the results of the research
%    project funded by the National Science Foundation.}

% \tnotetext[2]{The second title footnote which is a longer text matter
%    to fill through the whole text width and overflow into
%    another line in the footnotes area of the first page.}

\author[1]{Chandravardhan Singh Raghaw}%[type=editor,
                        % auid=000,bioid=1,
                        % prefix=Sir,
                        % role=Researcher,
                        % orcid=0000-0001-7511-2910
                        % ]

\cormark[1]


\ead{phd2201101016@iiti.ac.in}

% \ead[url]{www.cvr.cc, cvr@sayahna.org}

% \credit{Conceptualization of this study, Methodology, Writing - Original draft preparation}

\affiliation[1]{organization={Department of Computer Science and Engineering, Indian Institute of Technology Indore},
    addressline={Khandwa Road, Simrol}, 
    city={Indore},
    % citysep={}, % Uncomment if no comma needed between city and postcode
    postcode={453552}, 
    state={Madhya Pradesh},
    country={India}}

\author[2]{Arnav Sharma}%[style=chinese]
\ead{axs230011@utdallas.edu}


\author[1]{Shubhi Bansal}
\ead{phd2001201007@iiti.ac.in}

\author[1]{Mohammad Zia Ur Rehman}
\ead{phd2101201005@iiti.ac.in}

\author[1]{Nagendra Kumar}%[
   % role=Co-ordinator,
   % suffix=Jr,
\ead{nagendra@iiti.ac.in}
% \ead[URL]{www.sayahna.org}

% \credit{Mentorship}

\affiliation[2]{organization={Department of Computer Science, The University of Texas at Dallas},
    addressline={800 W Campbell Rd}, 
    city={Richardson},
    % citysep={}, % Uncomment if no comma needed between city and postcode
    postcode={75080}, 
    state={Texas},
    country={USA}}

% \author%
% [1,3]
% {Rishi T.}
% \cormark[2]
% \fnmark[1,3]
% \ead{rishi@stmdocs.in}
% \ead[URL]{www.stmdocs.in}

% \affiliation[3]{organization={STM Document Engineering Pvt Ltd.},
%     addressline={Mepukada}, 
%     city={Malayinkil},
%     % citysep={}, % Uncomment if no comma needed between city and postcode
%     postcode={695571}, 
%     state={Trivandrum},
%     country={India}}

\cortext[cor1]{Corresponding author: Chandravardhan Singh Raghaw}
% \cortext[cor2]{Principal corresponding author}

% \fntext[fn1]{This is the first author footnote. but is common to third
% \fntext[fn2]{Another author footnote, this is a very long footnote and
%   it should be a really long footnote. But this footnote is not yet
%   sufficiently long enough to make two lines of footnote text.}

% \nonumnote{This note has no numbers. In this work we demonstrate $a_b$
%   the formation Y\_1 of a new type of polariton on the interface
%   between a cuprous oxide slab and a polystyrene micro-sphere placed

\begin{abstract}
Swift and accurate blood smear analysis is an effective diagnostic method for leukemia and other hematological malignancies. However, manual leukocyte count and morphological evaluation using a microscope is time-consuming and prone to errors. Conventional image processing methods also exhibit limitations in differentiating cells due to the visual similarity between malignant and benign cell morphology. This limitation is further compounded by the skewed training data that hinders the extraction of reliable and pertinent features. In response to these challenges, we propose an optimized \textbf{Co}upled \textbf{T}ransformer \textbf{Co}nvolutional \textbf{Net}work (CoTCoNet) framework for the classification of leukemia, which employs a well-designed transformer integrated with a deep convolutional network to effectively capture comprehensive global features and scalable spatial patterns, enabling the identification of complex and large-scale hematological features. Further, the framework incorporates a graph-based feature reconstruction module to reveal the hidden or unobserved hard-to-see biological features of leukocyte cells and employs a Population-based Meta-Heuristic Algorithm for feature selection and optimization. To mitigate data imbalance issues, we employ a synthetic leukocyte generator. In the evaluation phase, we initially assess CoTCoNet on a dataset containing 16,982 annotated cells, and it achieves remarkable accuracy and F1-Score rates of 0.9894 and 0.9893, respectively. To broaden the generalizability of our model, we evaluate it across four publicly available diverse datasets, which include the aforementioned dataset. This evaluation demonstrates that our method outperforms current state-of-the-art approaches. We also incorporate an explainability approach in the form of feature visualization closely aligned with cell annotations to provide a deeper understanding of the framework.
\end{abstract}

% Coupled Transformer Convolutional Network (CoTCoNet)

% % Use if graphical abstract is present
% \includegraphics{CoTCoNet_2.png}

% % Research highlights
% \begin{highlights}
% \item An optimized novel coupled transformer-convolution network for leukemia detection
% \item Correlated global \& spatial features identify hematological malignancies efficiently
% \item Graph-based feature reconstruction module to identify hidden features of leukocytes
% \item A deep synthetic leukocyte generator is devised to mitigate data imbalance issues
% \item An extensive set of results verified on four diverse datasets (more than 16,982 WSIs)
% \end{highlights}

% Keywords

\maketitle

\section{Introduction}
Leukemia belongs to the broader class of blood cancer, wherein white blood cells (known as leukocytes) undergo malignant transformation into cancerous entities. These entities lead to an uncontrolled increase in the leukocytes that hampers average blood cell growth in leukemia. Regrettably, leukemia is the leading cause of cancer death worldwide, with a significant impact on global public health~\cite{Siegel2023}. Notably, the mortality rates tend to be disproportionately high in Low and Middle Income Countries (LMICs) due to the scarcity of early diagnosis and high-quality treatment, resulting in an increased morbidity rate. This results in the global burden of leukemia falling on LMICs, where 84\% of the cases are reported \cite{Asthana2018}.

Among the various subtypes of blood cancer, Acute Lymphocytic Leukemia (ALL) and Multiple Myeloma (MM) are most prominent. The term ``Acute'' in ALL signifies the rapid proliferation of immature White Blood Cell (WBC) blasts within the bone marrow. Conversely, MM is associated with diminished platelet counts in the blood. Treatment approaches for blood cancer hinge on several factors, including specific leukemia subtype, the age of the patient, rate of disease progression, and affected areas~\cite{Liu2019}. Among all, the blood cell count density plays an instrumental role in categorizing the precise subtype of blood cancer. Early diagnostic tests include blood cell count and morphological evaluation. The manual counting under the expertise of a skilled practitioner is a time-consuming process \cite{Kul2017} and necessitates expensive, sophisticated medical devices. Hence, there is a pressing need for automated computational methods that can overcome these limitations and omit the requirement of a skilled practitioner to run early diagnostic tests.

In recent years, several automated diagnostic methods for ALL have been introduced \cite{Mittal2022}. These methods depend on a predefined feature set designed to capture the cellular nucleus or cytoplasm for training classifiers used in ALL detection. However, a significant limitation of these methods is the utilization of small datasets to train the classifier, which can lead to overfitting and poor generalization of new blood smear samples. Moreover, the performance on smaller test sets may not be reliable \cite{Depto2023}. Another constraint for such techniques includes feature learning from raw data samples without prior segmentation of blood cells, which may not be optimal. These factors can hinder the effectiveness of these methods when deployed in real-world clinical settings. Several methodologies also incorporate local features by utilizing predefined or manually crafted features while neglecting the cohesive integration of spatial and global relationships among these features. As a consequence, these methodologies need to be more effectively capturing all pivotal features and associating them with classes in an efficient manner.

% \begin{figure*}[!b]
\begin{figure*}[ht]
  \centering
  \includegraphics[width=0.95\textwidth]{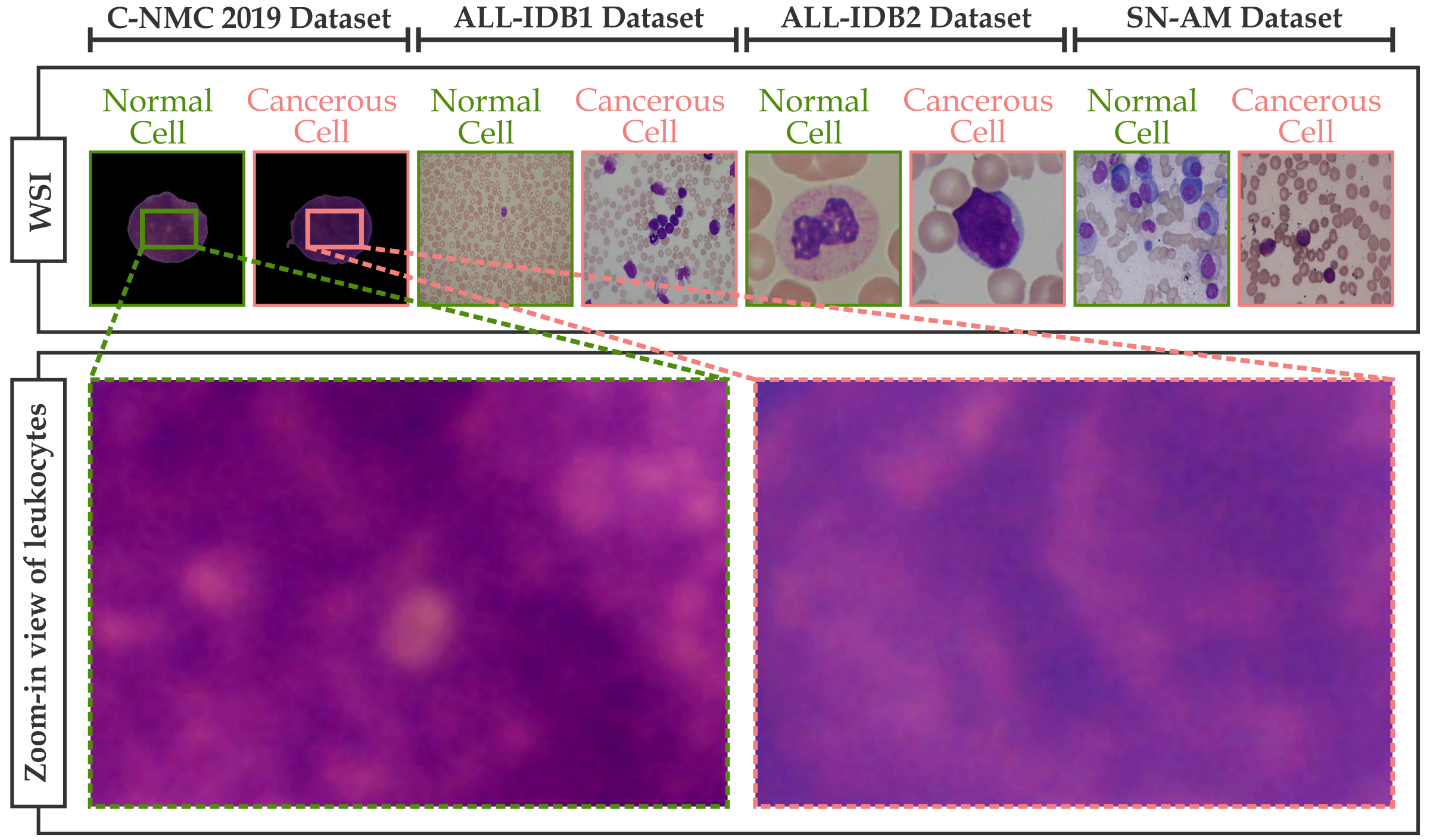}
  \caption{Illustration of sample images depicting the presence or absence of leukemic cells. The first row denotes the Whole Slide Image (WSI) of the blood samples, while the last row shows a zoomed-in view of WSI containing leukocytes. This figure highlights the indistinguishability of cell images across classes.}
  \label{fig:dataset-samples}
\end{figure*}

% Why existing approaches failed/drawbacks of previous papers
Recent advances in Machine Learning (ML) and Deep Learning (DL) have led to their successful application in medical diagnosis, including cancer survival \cite{Huang2023}, cognitive neuroscience \cite{Santamaria2023}, cardiology \cite{Langlais2023} and cancer image classification \cite{Jiang2023}. An essential prerequisite for accurate cancer image classification is precise segmentation since it enables a variety of quantitative analyses, such as shape, texture, and size. However, achieving accurate segmentation of pathology images proves to be complex due to factors such as blurred regions during image acquisition, noise interference, and low contrast between foreground and background elements. Furthermore, the segmentation process is compounded by significant variations in cell size, shape, and the heterogeneity of intracellular density \cite{Khandekar2021}. Among the various techniques employed for ALL detection, Convolutional Neural Networks (CNNs) have gained widespread popularity~\cite{Gehlot2020, Jawahar2022, Mohammed2023}. CNNs possess the capacity to extract task-specific local features directly from input blood smear images through a comprehensive learning approach. However, an inherent demand for a substantial volume of training data is required to achieve optimal performance. An alternative approach involves the utilization of transformers to capture global features of leukocytes \cite{Leng2023}. Despite this, morphological-based global features alone cannot accurately classify affected leukocytes. Furthermore, including high-dimensional features containing irrelevant, redundant, or noisy blood smear samples can lead to inaccurate diagnoses. Such feature vector tends to yield non-reproducible results and continues to exhibit considerable variance.

\autoref{fig:dataset-samples} shows the analysis of blood smears from the Whole Slide Image (WSI), illustrating the presence or absence of leukemic cells. The challenge arises from the fact that, in specific scenarios, peripheral blood smears exhibit visual characteristics that closely resemble one another, thereby complicating the diagnosis of leukemic cell presence. Leukocytes with similar visual features pose a challenge for feature extraction, including suboptimal staining quality, extensive overlapping of cell nuclei, and subtle morphological differences within and between leukocyte types. Currently, the detection of leukemic cells heavily relies on labor-intensive manual microscopy examinations, which not only consume a significant amount of time but also lack consistent objectivity~\cite{Heaven2022}. The complex and diverse leukocyte image features in the microscopic blood smear images pose a challenge in the development of a precise and efficient framework for the detection of leukemia~\cite{Varadarajan2023}. Hence, there is a need for a more efficient and generalized framework that can reduce time demands and enhance the accuracy of leukemia classification.

In light of the aforementioned challenges, we propose \textbf{Co}upled \textbf{T}ransformer \textbf{Co}nvolutional \textbf{Net}work (CoTCoNet) framework for efficient classification of leukocytes. CoTCoNet tackles the limited dataset challenges by generating synthetic leukocyte samples that closely mimic actual medical samples while preserving their inherent characteristics. It is imperative to note that these synthetic samples are meticulously designed to exclude any personally identifiable information, thereby ensuring the practical nullification of privacy concerns. Subsequently, CoTCoNet employs an exact segmentation technique to facilitate the accurate identification of the relevant regions of WBC within blood pathology images. In addition, a hybrid framework that capitalizes on the strengths of coupled transformers and convolutions is introduced to gain insight into the salient characteristics of blood smear samples. This framework comprehensively understands hematological features on both a global and spatial scale while also unveiling hidden features. Furthermore, we incorporate a feature selection technique to efficiently optimize the feature vectors, with the principal objective being the reduction of data dimensionality while preserving the most pertinent features. This objective aims to enhance the representational capacity of the selected features and the generalization capabilities of the framework. Finally, a classifier utilizes the optimized features to classify leukocytes precisely.

In summary, the primary objective of the research is to develop a framework that acquires knowledge of the hematological characteristics of blood cells to predict the presence or absence of leukemic cells within unseen leukocytes. These characteristics are acquired through extensive experiments on four diverse datasets. Experimental results demonstrate a notable superiority over existing methods. The key contributions of this work are summarized below:

\begin{itemize}
    \item We propose a novel framework, the Coupled Transformer Convolution Network (CoTCoNet), designed to classify leukemia cells. CoTCoNet integrates the Global Feature Module (GMod) and the Spatial Feature Module (SMod) to combine long-range contextual information and scalable spatial patterns, identifying complex and large-scale hematological features.
    \item The proposed Graph-based Feature Reconstruction (GraFR) module is designed to unveil hidden features by calculating similarity with prominent neighbor features while preserving the geometric information of the blood cells.
    \item We employ an enhanced Population-based Meta-Heuristic Algorithm for feature selection and optimization, aiming for a harmonious balance between the influential feature mapping in the negative and positive classes of leukocyte cells.
    \item We introduce LeuGAN, an architecture powered by Generative Adversarial Networks that effectively manages class imbalance by generating high-resolution leukocyte samples. Simultaneously, it preserves the extensive global and spatial information found within blood cell images by utilizing an interconnected feedback loop.
    \item CoTCoNet framework effectively utilizes the finely-tuned Leukemia Segment Anything Model (LeuSAM) to segment leukocyte cells. LeuSAM proficiently identifies atypically shaped blood cells with limited surface areas, effectively excluding extraneous regions, such as plasma, and precisely segments the primary area for more efficient feature extraction.
    \item We conducted comprehensive experiments and in-depth analyses on four distinct publicly available datasets to demonstrate the effectiveness of the CoTCoNet framework. The superior performance of the model, when compared to existing methods for leuke-mia detection, validates its robustness.
\end{itemize}

This article is organized into distinct sections, as follows. \autoref{sec:related} provides an in-depth examination of existing literature on leukemia classification. \autoref{sec:methodology} provides an elaborate account of the proposed methodology. \autoref{sec:exp-eval} provides an overview of the evaluation process conducted during the experiments. Lastly, \autoref{sec:conclusion} provides the primary findings, followed by examining prospective avenues for future research.

\section{Related Work}
\label{sec:related}
This section provides a comprehensive survey of the previous literature on leukemia classification. For better understanding, we divide the leukemia classification techniques into three categories: Deep Learning-based Leukemia Detection, Segmentation-based Leukemia Detection, and Class Imbalance in Leukemia Detection.

\subsection{Deep Learning-based Leukemia Detection}
Deep Learning (DL) methods have achieved promising results in leukemia cell classification from microscopic blood smear images in recent studies~\cite{Abhishek2023, Wang2023, Jiwani2023}. DL methods learn features and patterns autonomously, leading to effective leukemic cell classification. Mohammed~\textit{et~al.} introduced a stacked ensemble framework~\cite{Mohammed2023}, which harnessed transfer learning techniques, specifically ResNet-101, GoogleNet, SqueezeNet, DenseNet-201, and Mobile-NetV2, for leukemia detection. In contrast, CoTCoNet capitalizes on the capability of coupled transformers and convolution filters to learn long-range dependencies integrated with local features. Gehlot~\textit{et~al.}~\cite{Gehlot2020} proposed SDCT-AuxNet$\theta$, which focuses on stain absorption and sparse features. SDCT-AuxNet$\theta$ incorporates a base CNN integrated with bilinear pooling and the Stain-Deconvolution (SD)-Net architecture to improve features with Optical Density-space analysis. Jawahar~\textit{et~al.}~\cite{Jawahar2022} designed a neural network featuring three cluster layers that employ a combination of convolution and max-pooling operations to extract hierarchical and robust features. The features generated have spatial information but are not fully optimized. CoTCoNet, however, leverages a population-based algorithm for feature optimization to enhance the model's ability to extract and learn complex patterns in the data.

Das~\textit{et~al.} proposed a novel Orthogonal SoftMax Layer (OSL)-based model~\cite{Das2022} that combines ResNet 18-based feature extraction and OSL-based classification. OSL improves classification by making weight vectors orthogonal. CoTCoNet, on the other hand, incorporates a Hierarchical Deep Learning Classifier (HDLC) that can classify effectively even when the cell images exhibit striking visual similarity. Dhalla~\textit{et~al.} introduced an encoder-decoder architecture~\cite{Dhalla2023} that prioritizes extracting local multiscale features. The attention module in the architecture filters the most relevant features, focusing on spatial attributes. Das~\textit{et~al.} utilized a transfer-learning-based approach by employing MobileNetV2~\cite{Sandler2018} for feature extraction, enhanced with depthwise separable convolution for computational efficiency. Atteia~\textit{et~al.} presented a novel Bayesian-based optimized CNN~\cite{Atteia2022} that employs an informed iterative procedure to explore the hyperparameter configuration that yields the best network settings for minimizing an objective error function. In contrast, the CoTCoNet framework utilizes a graph-based feature reconstruction technique to identify hidden features. Subsequently, it leverages a population-based meta-heuristic optimization technique to derive optimized reconstructed features, enabling efficient classification.

\subsection{Segmentation-based Leukemia Detection}
Medical image segmentation is essential for the analysis and diagnosis of diseases, as it allows for the precise identification of Regions of Interest (ROIs)~\cite{Hoorali2022, Qureshi2023} within samples. Devi~\textit{et~al.}~\cite{Devi2023} employed Gaussian Blur for image denoising and applied morphological operations to create an image mask. Subsequently, binary thresholding was applied to the mask to generate cell contours, enabling the segmentation and identification of malignant cells. However, the CoTCoNet framework adeptly leverages the finely-tuned Leukemia Segment Anything Model (LeuSAM) to accurately identify irregularly shaped blood cells with limited surface areas. Khandekar~\textit{et~al.}~\cite{Khandekar2021} leveraged the You Only Look Once (YOLOv4) architecture to detect unhealthy White Blood Cells (WBC). YOLOv4 incorporates cross-stage-partial connections for the precise prediction of bounding boxes within the designated ROIs. Alagu~\textit{et~al.}~\cite{Alagu2021} utilized the UNet architecture to extract in-depth features from models like AlexNet, GoogleNet, and SqueezeNet, which enabled accurate segmentation and discrimination of healthy and blast cell nuclei. 

Mohammed~\textit{et~al.}~\cite{Mohammed2021} proposed a technique for segmenting WBC and deriving relevant features from the segmented cells. This method employs threshold segmentation, which leads to the generation of segmented WBCs. Subsequently, the segmented WBCs undergo a discrete cosine transformation, and the resulting coefficients serve as classification features. In contrast, CoTCoNet extracted combined global and spatial features from transformer and convolutional-based modules for efficacious classification. Das~\textit{et~al.}~\cite{Das2020detection} segmented the lymphocyte by employing hand-crafted features involving a color-based k-means clustering technique. Additionally, a gray-level co-occurrence matrix and gray-level run-length matrix extract the relevant features from lymphocytes. Das~\textit{et~al.} proposed a hybrid ellipse fitting-based blood-cell segmentation~\cite{Das2022efficient} mechanism utilizing a Laplacian-of-Gaussian (LoG) and leverages fast radial symmetry-based seed-point detection to precisely segment overlapping cells and low-contrast visual features. On the other hand, the CoTCoNet framework enhances the visibility of low-contrast features by implementing a contrast-limited adaptive histogram equalization technique. Furthermore, the finely tuned LeuSAM excludes extraneous regions, such as plasma, and achieves precise segmentation, thereby enhancing the efficiency of feature extraction.

\subsection{Class Imbalance in Leukemia Detection}
Many real-world applications, such as fraud detection \cite{Wang2020}, intrusion identification~\cite{Panigrahi2019}, and cancer diagnosis~\cite{Saxena2021}, face classification problems with imbalanced data distribution~\cite{Farshidvard2023}. Leukemia detection is very challenging due to the large imbalance in the number of samples between the minority and majority classes. Chen~\textit{et~al.} proposed the WBC-GLAformer~\cite{Chen2023}, a hybrid transformer-based model to acquire local and global features. The WBC-GLAformer comprises a low-level feature extractor to extract local features and a global-local attention encoder to combine the strengths of local and global features. However, CoTCoNet integrates spatial and global features and incorporates hidden features through graph-based feature reconstruction.
Umar~\textit{et~al.}~\cite{Umer2023} trained a robust CNN by addressing cross-domain data imbalance, domain shifts, and missing classes. The author shows an improvement in existing pre-trained deep models by optimizing the loss function in the network. Depto~\textit{et~al.}~\cite{Depto2023} highlighted the imbalance class problem and presented detailed quantitative and qualitative analyses for leukemia classification. The result demonstrates that Generative Adversarial Networks and loss-based methods efficiently mitigate class imbalance challenges. However, CoTCoNet utilizes fine-tuned LeuGAN by synthesizing high-quality leukocyte samples, preserving the morphological information by leveraging an inter-connected feedback loop.

\section{Methodology}
\label{sec:methodology}
\begin{figure*}
  \centering
  \includegraphics[width=0.95\textwidth]{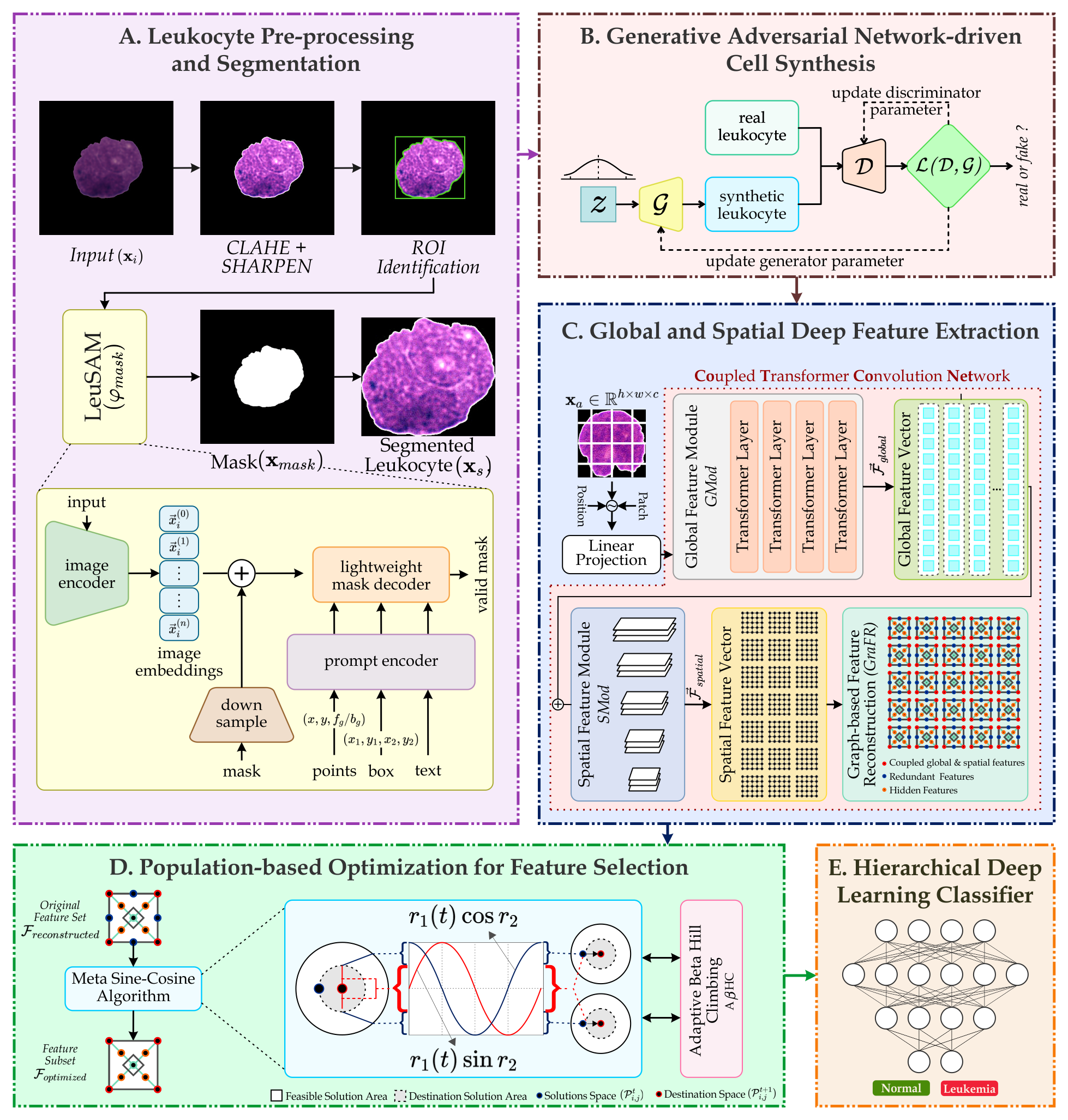}
  % \caption{Proposed CoTCoNet}
  \caption{Proposed Coupled Transformer Convolution Network (CoTCoNet) for Leukemia Detection. Firstly, the leukocyte cell images are inputted into \textcolor{blue}{(A)} for pre-processing and segmentation. In step \textcolor{blue}{(A)}, the fusion of CLAHE and Sharpen image enhancement technique is applied, followed by Region of Interest Identification. Then, images are segmented by utilizing LeuSAM. The output of Step \textcolor{blue}{(A)} is fed to the proposed GAN-driven cell synthesis step \textcolor{blue}{(B)}. The outcome of step \textcolor{blue}{(B)} comprises synthetic cell images produced by the GAN adhering to the real images. In step \textcolor{blue}{(C)}, CoTCoNet first isolates the global and spatial features and then proceeds to regenerate them through graph-based feature reconstruction techniques. At step \textcolor{blue}{(D)}, a meta-heuristic sine-cosine algorithm optimizes the feature. Finally, step \textcolor{blue}{(E)} classifies the optimized features into normal or leukemia. The effectiveness of the proposed framework is assessed on four different datasets.}
  \label{fig:cotconet}
\end{figure*}

This section includes a comprehensive breakdown of the proposed \textbf{Co}upled \textbf{T}ransformer \textbf{Co}nvolution \textbf{Net}work (CoTCoNet) framework, as illustrated in~\autoref{fig:cotconet}, which comprises five distinct sub-modules: (A) leukocyte pre-processing and segmentation undertakes the initial processing and segmentation of cell images; (B) generative adversarial network-driven cell synthesis responsible for generating synthetic cells; (C) global and spatial deep feature extraction extracts integrated global and spatial features, coupled with the identification of hidden features; (D) population-based optimization for feature selection selects and refines the most optimized set of features; (E) hierarchical deep learning classifier categorizes cells as either normal or leukemic.

% \begin{figure*}[!ht]

\subsection{Problem Formulation}
\label{sec:formulation}
This work proposes an advanced framework for acute lymphoblastic leukemia detection. It accurately identifies blast or leukemic cells in the blood, extracting their unique features through global and spatial modules. Given a set of leukocyte cell samples $\mathscr{X} \in \{x_1,x_2,\cdots,x_n \}$ and the corresponding labels $\mathscr{Y}$, we aim to predict the presence or absence of leukemic cells in unseen leukocyte blood cell images. The prediction is based on the feature learning framework, which learns the hematological features of cells in a supervised manner and creates the custom mapping $f\!:\!\mathscr{X}\! \rightarrow\! \mathscr{Y}$, such that $\mathscr{Y}\in (0,1)$ and the goal is to predict the presence or absence of leukemic cells in a blood cell image.

\subsection{Data Preprocessing}
\label{sec:preprocessing}
We devise a leukocyte cell pre-processing and segmentation pipeline, as illustrated in \autoref{fig:cotconet}. Firstly, we employ Contrast Limited Adaptive Histogram Equalization (CLAHE) for image contrast amplification and visibility enhancement. After that, we sharpen filters to emphasize the edges and features of the microscopic images. Secondly, we identify and capture Region-of-Interest (ROIs) that are more likely to exhibit variations indicative of leukemia while excluding any superfluous sections such as plasma. Subsequently, we leverage the fine-tuned Leukemia Segment Anything Model (LeuSAM)~\cite{Kirillov2023} to perform segmentation and extraction of leukocyte cells from the ROI images, resulting in the production of segmented leukocytes $\textbf{x}_s$. Finally, the array of segmented images is utilized as input for a mechanism driven by a Generative Adversarial Network, with the aim of synthesizing cells.

\subsection{Generative Adversarial Network-driven Cell Synthesis}
\begin{figure}[!ht]
  \centering
  \includegraphics[width=\columnwidth]{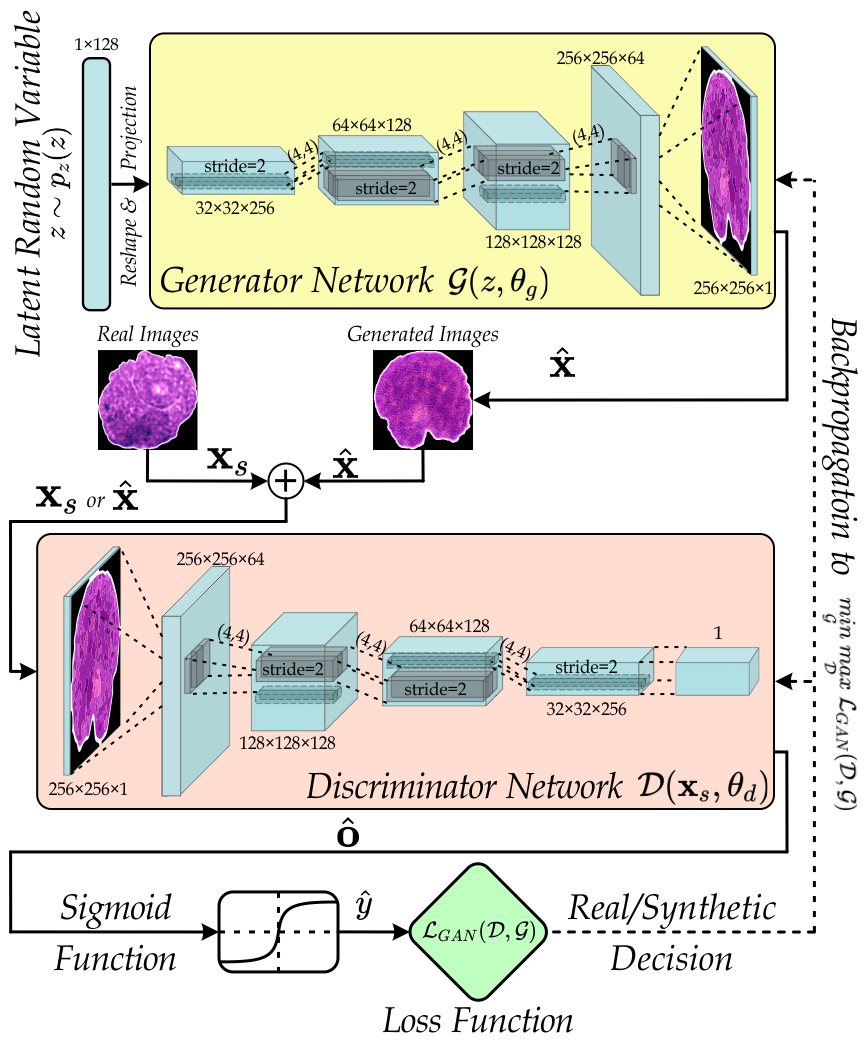}
  \caption{Proposed Generative Adversarial Network-driven architecture (LeuGAN) for synthetic leukocyte cell generation. Our approach involves a $128$-dimensional noise vector $z$, Generator $\mathcal{G}$ and corresponding Discriminator $\mathcal{D}$. Specifically, $\mathcal{G}$ uses the noise vector to craft synthetic leukocyte blood cells, while $\mathcal{D}$ evaluates both real and generated leukocytes combined. The network employs the generator-adversarial loss function $\mathcal{L}_{GAN}(\mathcal{D,G})$ to maintain consistency effectively.
  }
  \label{fig:leugan}
\end{figure}

This section introduces LeuGAN, a novel method for generating synthetic hematological cells while preserving essential features. LeuGAN consists of a twin neural network, a generator $\mathcal{G}$ and one discriminator $\mathcal{D}$. The trained $\mathcal{G}$ effectively characterizes a probability distribution $p_z(z)$ in the training of segmented leukocyte $\textbf{x}_s$. The generator devises synthetic mapping $\mathcal{G}(\textbf{z},\theta_g)$ from $p_z(\textbf{z})$, where $\theta_g$ represents the learning parameter of the generator network. As illustrated in ~\autoref{fig:leugan}, $\textbf{z}$ is $128$ dimensional vector with each $z_i\in (-1,+1)$ is transformed in $1024$ feature maps of distributed convolutions. In order to synthesize high-fidelity artificial leukocytes, a sequence of four strided convolutions is employed to project and reshape the noise vector $\textbf{z}$ into a $256\times256$ pixel representation of leukocyte cell image, denoted as $\hat{\textbf{x}}$.

The Discriminator Network $\mathcal{D}(\textbf{x}_s,\theta_d)$, where $\theta_d$ denotes the discriminator learning parameter, accepts either a synthesized leukocyte $\hat{\textbf{x}}$ or real leukocyte $\textbf{x}_s$ cell image. As shown in~\autoref{fig:leugan}, inputs traverse from four convolution layers and generate an output $\hat{o}$ which indicates whether the input image is real or synthetic; it can be mathematically represented in \autoref{eq:sigmoid} as follows:

 \begin{equation}
\label{eq:sigmoid}
    \hat{y} =  \frac{e^{\hat{o}}}{1 + e^{\hat{o}}} \quad \textit{s.t.}\quad\hat{y} \in [0,1]
\end{equation} 
 
where $\hat{y}$ denotes the class = \{synthetic leukocyte, real leukocyte\}. The final objective is to increase the classification performance of real leukocytes and synthetic leukocyte cells, with the generator network $\mathcal{G}(\textbf{z},\theta_g)$ and discriminator network $\mathcal{D}(\textbf{x}_s,\theta_d)$ improves learning by incorporating the classification loss. The adversarial interplay between $\mathcal{G}$ and $\mathcal{D}$ integrates the binary loss and can be depicted in \autoref{eq:lossgan}:

 \begin{equation}
    \begin{aligned}
    \label{eq:lossgan}
    \underset{\mathcal{G}}{min}\,\underset{\mathcal{D}}{max}\,\mathcal{L}_{GAN}(\mathcal{D},\mathcal{G})= \\
    & \hspace{-33.5mm}\underset{\textbf{x}_s\sim\textbf{p}_{data}}{\mathbb{E}}(\textbf{x}_s)\left[log\bigl(\mathcal{D}(\textbf{x}_s)\bigr)\right]+
    \underset{\textbf{z}\sim\textbf{p}_z(z)}{\mathbb{E}}(\textbf{x}_s)\left[log\Bigl(1-\mathcal{D}\bigl( \mathcal{G}(\textbf{z}) \bigr)\Bigr)\right]
    \end{aligned}
\end{equation}  

where the generator aims to minimize the adversarial loss $\mathcal{L}_{GAN}(\mathcal{D},\mathcal{G})$, while the discriminator strives to maximize it. To train the discriminator network, the generator $\mathcal{G}$ works in backpropagation (i.e., in an interlinked feedback loop), and to train the generator network, the discriminator does the same. After generating the synthetic leukocyte cells, we introduce them into the feature extraction module integrated within the provided framework. We elucidate this process in the following~\autoref{sec:cotconet}. 

\subsection{Coupled Transformer Convolution Network (CoTCoNet)}
\label{sec:cotconet}
This section unveils the arranged configuration of the CoTCoNet framework, as illustrated in~\autoref{fig:cotconet}. The CotCoNet comprises three integral modules: Global Feature Module (GMod), Spatial Feature Module (SMod), and Graph-based Feature Reconstruction module. This framework involves leveraging a well-designed transformer-based GMod to capture the overarching global features effectively. At the same time, the SMod module capitalizes on the extraction capabilities of the Convolutional Neural Network (CNN) by inheriting the spatial features. Subsequently, the combined features undergo a reconstruction process utilizing a Graph-based Feature Reconstruction (GraFR) module to extract the pertinent features of each data point by evaluating the similarity score of feature vectors. These three modules form a cohesive network that collaboratively leverages the extracted features.

\subsubsection{Global Feature Module (GMod)}
\label{sec:gmod}
\begin{figure}[!ht]
  \centering
  \includegraphics[width=\columnwidth]{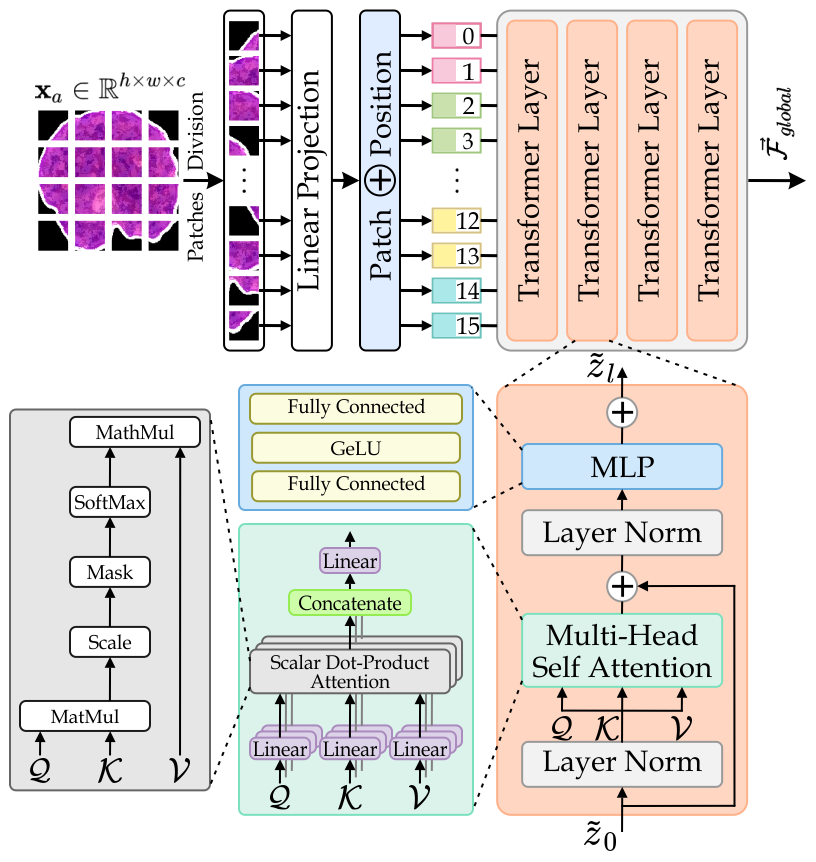}
  \caption{Architecture of the proposed Global Feature Module \textit{(top)} and details of the Transformer Layer block \textit{(bottom)}. The input image is first split into equal-sized patches and flattened. Then, each patch is projected into a feature space where a transformer layer block processes them to extract global features.}
  \label{fig:gmod}
\end{figure}

We design a transformer-based network called Global Feature Module (GMod) that focuses on extracting global information from the input images to classify leukemic cells accurately. The GMod module is schematically illustrated in \autoref{fig:gmod}.

\paragraph{\textbf{Patch Decomposition}}
The standard transformer accepts a 1-D sequence of token embeddings as its primary input; however, the manipulation of 2-D synthetic leukocyte images involves a reshaping operation of input $\textbf{x}_a \in \mathbb{R}^{h \, \times \, w \, \times \, c}$ to $\textbf{x}_p \in \mathbb{R}^{n \, \times \, (p^2 \, \cdot \, c)}$, where $(h,w,c)$ corresponds to height, width, channels and $(p, p)$ indicates the spatial resolution of an individual patch. The overall count of these patches, referred to as $n$, is computed as $(h \times w) / p^2$, furnished as input to the stacked transformer layers. 

\paragraph{\textbf{Linear Transformation and Positional Embedding}}
We apply the trainable linear projection to map the vectorized token patches $\textbf{x}_p$ into a latent embedding space with $d_{L}$ dimensionality. The token sequence denoted as $\textbf{x}_p =[\textbf{x}_p^{(1)},\textbf{x}_p^{(2)} \cdots \textbf{x}_p^{(N)}]$ is initially structured as an array after incorporating the learnable embedding and the position embedding the resultant $\Tilde{z}_0$, [refer~\autoref{eq:positional-embed}] is directed towards the transformer, whose output is given by $\Tilde{z}_L$, where $L$ signifies the number of transformer blocks. Subsequently, we integrate position embeddings into token embedding to inherit the positional information as follows:

\begin{equation}
\label{eq:positional-embed}
    \Tilde{z}_0 =  [\textbf{x}_p^{(1)}\textbf{E} ; \: \textbf{x}_p^{(2)}\textbf{E} ; \: \cdots ; \: \textbf{x}_p^{(N)}\textbf{E}] + \textbf{E}_{pos}
\end{equation}

where, $\textbf{E} \in \mathbb{R}^{(p^2 \, \cdot \, c) \, \times \, d_L}$ denotes the patch embedding projection, and $\textbf{E}_{pos}\in \mathbb{R}^{(n+1)\, \times \, d_L}$ represents the positional embedding. 

\paragraph{\textbf{Self-Attention Mechanism}}
The resulting token embeddings $\Tilde{z}_0$ are introduced as input to a transformer that consists of a Multihead Self-Attention (MSA) blocks and Multi-Layer Perceptrons (MLP) blocks [refer~\autoref{eq:zmsa} and~\autoref{eq:zmlp}]. The MLP block includes a dual fully connected dense layer with a Gaussian error Linear Unit (GeLU) residing between them. The Layer Normalization (LN), as defined in~\autoref{eq:zln}, is employed prior to the MSA and MLP blocks, while residual concatenation is utilized after each block. The procedure for the forward calculation is outlined as follows:

\begin{equation}
\label{eq:zmsa}
    \Tilde{z}_l=M\!S\!A\bigl( L\!N(z_{l-1}) \bigr) + z_{l-1}, \quad l=1, \ldots , L
\end{equation}
\vspace{-7mm}
\begin{equation}
\label{eq:zmlp}
    z_l=M\!L\!P\bigl( L\!N(\Tilde{z}_l) \bigr) + \Tilde{z}_l, \quad l=1, \ldots , L
\end{equation}
\vspace{-7mm}
\begin{equation}
\label{eq:zln}
    y=L\!N(z_L^0)
\end{equation}

The core mechanism of the transformer is MSA, which aims to capture correlated global contextual information effectively. To utilize diverse implications, three adaptable weight matrices ($\textbf{W}^{\mathcal{Q}}, \, \textbf{W}^{\mathcal{K}}, \, \textbf{W}^{\mathcal{V}}$) are established. Simultaneously, $\textbf{W}^{\mathcal{Q}} \in \mathbb{R}^{h \times w \times c}, \textbf{W}^{\mathcal{K}} \in \mathbb{R}^{h \times w \times c}, \textbf{W}^{\mathcal{V}} \in \mathbb{R}^{h \times w \times c}$ are initialized randomly. Leveraging these trainable weight matrices facilitates a linear transformation of tokens into a 3-D invariant matrix comprising Queries ($\mathcal{Q} \in \mathbb{R}^{h \times w \times c}$), Keys ($\mathcal{K} \in \mathbb{R}^{h \times w \times c}$), and Values ($\mathcal{V} \in \mathbb{R}^{h \times w \times c}$). Subsequently, the computation of attention scores involves the entire information encompassed within $\mathcal{Q}, \, \mathcal{K}, \, \mathcal{V},$ and the weights are calculated using the softmax function. Concisely, the SA mechanism can be encapsulated in \autoref{eq:sa}:

 \begin{equation}
\label{eq:sa}
    S\!A = Attention (\mathcal{Q}, \, \mathcal{K}, \, \mathcal{V}) = softmax \left( \frac{\mathcal{Q}\mathcal{K}^T}{\sqrt{d_{\mathcal{K}}}} \right) \mathcal{V}
\end{equation} 

where, $d_{\mathcal{K}}$ is the dimension of $\mathcal{K}$. The MSA module is equipped with multiple weight matrices to facilitate the mapping of $\mathcal{Q}$, $\mathcal{K}$, and $\mathcal{V}$ components; this plays a crucial role in computing MSA values through uniform operational procedures. Then, the outcomes from individual attention heads are combined through concatenation. The linear transformation represented in \autoref{eq:msa} is the final step in deriving the MSA outcomes.

 \begin{equation}
\label{eq:msa}
    M\!S\!A(\mathcal{Q}, \mathcal{K},\mathcal{V})=Concat(S\!A_1, S\!A_2, \dots , S\!A_H)\textbf{W}
\end{equation} 

where, $H$ represent total number of heads, $\textbf{W}\in \mathbb{R}^{h \, \times \, d_{\mathcal{K}} \, \times \, d_{\mathcal{W}}}$ denotes the parameter matrix of he linear layers and $d_{\mathcal{W}}=N+1$ corresponds to token number.

The pre-trained MSA module generates the feature matrix and is directed into the MLP module.
Also, the MLP is followed by Layer Normalization (LN) to amplify gradient magnitudes, mitigating the gradient vanishing concerns and accelerating the training process. In the transformer module, the dimensions of the input $z_{l-1}$ and the output $\Tilde{z}_{\mathcal{l}}$ from the $l^{th}$ transformer module are identical. The trainable embedding $z_L^0$, comprising a set of global leukocyte features $\vec{\mathcal{F}}_{global}$ is fed into the Spatial Module consisting of several convolutions for capturing spatial features.

\begin{figure*}[!ht]
  \centering
  \includegraphics[width=\textwidth]{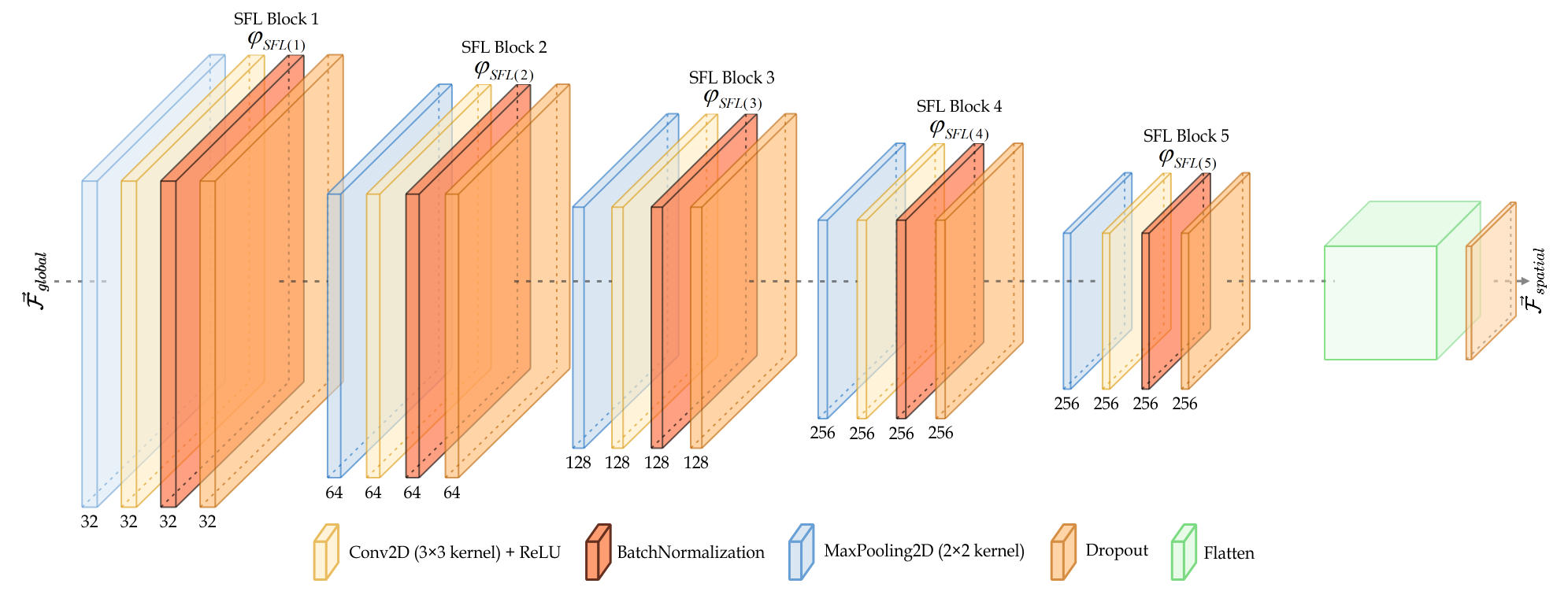}
  \caption{ Architecture of the proposed Spatial Feature Module, comprising five Spatial Feature Learning (SFL) blocks, represented by $\varphi_{S\!F\!L(i)}$, $i \in \left[1,2,3,4,5 \right]$. The numerical values at the bottom indicate the number of filters employed in each SFL block.}
  \label{fig:smod}
\end{figure*}

\subsubsection{Spatial Feature Module (SMod)}
\label{sec:smod}
Transformer-based GMod can capture global features efficiently, but it is less effective than Convolution Neural Network (CNN) to learn scalable spatial patterns. Hence, a Convolutional-based Spatial Module (SMod) is incorporated to learn spatial features at different scales efficiently. As presented in ~\autoref{fig:smod}, SMod architecture consists of five Spatial Feature Learning (SFL) blocks, denoted as $\varphi_{S\!F\!L(i)}$, $i \in \left[1,2,3,4,5 \right]$. Using this concurrent structure, the system can extract and handle spatial information while maintaining the integrity of global information as much as possible. SMod is defined as a dual feature extraction block as it extracts the spatial features from the global features of GMod. 

As shown in~\autoref{fig:smod}, the SFL block begins with a cascaded arrangement of convolution layer \textit{(CONV)}, batch normalization layer \textit{(BN)}, max-pooling layer \textit{(MXP)}, \textit{ReLU} activation function and dropout layer \textit{(DRP)}. The \textit{CONV} layer employs a spatial filter with dimensions $3 \times 3$ to capture the detailed features at the finest levels from global features. Notably, the padding configuration is set to ``same'' to preserve the homogeneity in feature map dimensions throughout the SFL block. The presence of the \textit{BN} layer mitigates overfitting and accelerates the learning process. Incorporating a \textit{ReLU} activation function introduces non-linearity during the learning process. A dropout layer is parameterized by a value of $0.2$, a regularization to address the overfitting. Consequently, this layer selectively randomly deactivates neurons to reduce the dependency on any single neuron and leads to more robust feature learning.

Let $\varphi_{S\!F\!L(i)}^f(\Ddot{\mathcal{C}})$ represent the output of a SFL block with $f$ filters, $\varkappa_s^{\tilde{k},\tilde{k}}$ denotes the \textit{MXP} with kernel size $\tilde{k} \times \tilde{k}$ and stride $s$. $\Psi^{k,k,f}_{S\!F\!L(i)}(\Ddot{\mathcal{C}})$ represent convolution operation with $f$ number of kernels of size $k \times k$, $\beta$ denote the \textit{BN}, and $\rho$ denote the \textit{ReLU} activation function. \autoref{eq:sflblock} mathematically formalizes the functioning of the SFL block $\varphi_{S\!F\!L(i)}^f(\Ddot{\mathcal{C}})$, as outlined below:

 \begin{equation}
\label{eq:sflblock}
   \varphi_{S\!F\!L(i)}^f(\Ddot{\mathcal{C}})\; =\; 
   \varkappa_2^{2,2}
   \left( \rho \left( \beta \left( \Psi^{3,3,f}(\Ddot{\mathcal{C}}) \right) \right) \right)
\end{equation} 

In the SFL block, the initial phase involves channeling the input through a max-pooling layer featuring a kernel size of $2 \times 2$ with stride value $2$ to reduce the size of the feature map. Subsequently, the resultant is fed into a convolution \textit{CONV} layer employing a $3 \times 3$ kernel size and incorporating distinct filters denoted as $f_i$, where $i \! \in \! [32,64,128,256,256]$ represents the filter values of each SFL block. As a result, the procedure generates a tensor of spatial feature maps consisting of the fine-grained features of the leukocyte cell, such as textures, corners, edges, and colors. Next, the generated feature maps are fed into \textit{BN} to elevate the feature learning mechanism by normalizing the input through the computation of the mean and variance within the input mini-batch. This facilitates managing alterations to the input distribution layer by changing vital parameters of preceding layers. Consequently, this operation enhances the speed of convergence as well as the generalization ability of the network. Lastly, a dropout layer (\textit{DRP}) is introduced to mitigate the co-adaptation of features, thereby compelling the network to develop more robust representations. Finally, the SFL blocks generate a flattened output containing spatial features $\vec{\mathcal{F}}_{spatial}$, which is then fed into the Graph-based feature reconstruction (GraFR). GraFR aims to improve the quality of the features by utilizing aggregated information from all preceding layers within the SFL blocks, thereby enhancing the feature reconstruction. This reconstruction improves the feature quality, which measures how well the features represent the leukocyte cell and its type.

\subsubsection{Graph-based Feature Reconstruction (GraFR)}
\label{sec:grafr}
The complex and hard-to-see biological features of the leukocyte cell are challenging to emphasize. This requires a method focused on reconstructing hidden or unobserved features from a lower-dimensional to a higher-dimensional feature map. Hence, a Graph-based feature reconstruction (GraFR) is proposed to scale up the unobserved sample features. The primary aim behind this technique is to amplify the hidden features of nodes by leveraging the features of their neighboring nodes. The GraFR module establishes a connection between feature elements using low-level feature maps to steer the up-sampling process using high-level feature maps. GraFR considers the whole feature information, bridging the gap between global and local perspectives of semantic information. The GraFR module follows four steps: (1) Node generation for feature maps, (2) Edge weight computation, (3) Hidden feature selection criteria, and (4) Feature map reconstruction.

\paragraph{\textbf{Node Generation for Feature Maps}}
Let $\mathbb{G}(V_g,E_g)$ be the graph, where $V_g$ is the set of nodes, and $E_g$ is the set of edges. Each node $v$ in the graph is associated with a one-dimensional (1D) flattened array $x_v$ that contains coupled global and spatial features. This can be represented as $\vec{x_v} = \left[ x_{v,1}, x_{v,2}, \cdots , x_{v,d} \right]$ where $d$ represents the dimensionality of the feature vectors.

\paragraph{\textbf{Edge Weight Computation}}
The edge weights in the graph are calculated using the Euclidean distance between the feature vectors of nodes. This means that for any two nodes $u$ and $v$ in $V_g$, the Euclidean distance $d_{uv}$ between their feature vectors $x_u$ and $x_v$ is given in \autoref{eq:euclidist}:

 \begin{equation}
\label{eq:euclidist}
   d_{uv}(u, v)=\sqrt{\sum_{i=1}^n \| \vec{x}_{u_i} - \vec{x}_{v_i} \|^2}
\end{equation} 
 
 where, $\vec{x}_{u_i}$ and $\vec{x}_{u_i}$ are the $i^{th}$ elements of $x_u$ and $x_v$, respectively.

\paragraph{\textbf{Hidden Feature Selection Criteria}}
The criteria for selecting the hidden features $\hbar$ aims to identify nodes with the highest similarity scores. The node incorporates global and spatial features while considering all features from lower-dimensional representations. The similarity score $\nabla_{u \leftrightarrow v}$ between nodes $u$ and $v$ is the inverse of the edge weight, as shown in \autoref{eq:similarityscore}:

 \begin{equation}
\label{eq:similarityscore}
   \nabla_{u \leftrightarrow v}=\left( \frac{1}{d_{uv}} \right)
\end{equation} 

A higher similarity score indicates a stronger similarity between core and hidden features. For inclusion of the feature node, each node $i$ in $V_g$ is considered if it possesses the highest similarity score across all other nodes in $V_g$, as shown in the following~\autoref{eq:sscriteria},

 \begin{equation}
\label{eq:sscriteria}
   \hbar=\{i\in V_g\:|\:\forall v \in V_g,\: (\nabla_{i \leftrightarrow v} \geq \nabla_{v \leftrightarrow i}) \}
\end{equation} 

The set of hidden features $\hbar$ includes node $i$ if this condition is satisfied.

\paragraph{\textbf{Feature Map Reconstruction}}
The output of the Feature Map Reconstruction is a vector $\mathcal{F}_{reconstructed}$ that summarizes all features and highlights hidden features of the nodes in the graph. This can be obtained by weighing the contribution of each node's feature vector based on the similarity score. \autoref{eq:freconstruction} is used to compute the reconstructed features $\mathcal{F}_{reconstructed}$.

 \begin{equation}
\label{eq:freconstruction}
   \mathcal{F}_{reconstructed}=\frac{\sum_{v=V_g} \, \nabla_{u \leftrightarrow v} \cdot x_v}{\sum_{v=V_g} \, \nabla_{u \leftrightarrow v}}
\end{equation} 

This reconstructed feature $\mathcal{F}_{reconstructed}$ incorporates information from all nodes in the graph, with nodes that have higher similarity scores contributing more significantly to the final representation of reconstructed features. CoTCoNet framework procedure is outlined in \autoref{alg:cotconet}. We extract global features from Lines 1-5 and spatial features from Lines 6-12. Lines 13-16 perform graph-based feature reconstruction to identify hidden features.

\begin{figure*} % Full-width Algorithm
\centering{
\begin{minipage}{0.9\linewidth}
  \begin{algorithm}[H]
    \caption{\textsc{CoTCoNet Framework}}
    \label{alg:cotconet}
    \begin{algorithmic}
        \vspace{1mm}
        \STATE
        \textbf{Input:} A set of augmented images $\textbf{x}_a$ of size $\mathbb{R}^{h \, \times \, w \, \times \, c}$ \\
        \vspace{0.8mm}
        \textbf{Output:} Reconstructed features $\vec{\mathcal{F}}_{reconstructed}$ \\
        \vspace{1mm}
        \textbf{procedure:} \textit{GMod} $\bigr( \textbf{x}_a \in \mathbb{R}^{h \, \times \, w \, \times \, c},\, \vec{\mathcal{F}}_{global}  \bigr)$   
    \end{algorithmic}
    \begin{algorithmic}[1]
    \STATE Decompose images into patches with no overlap. \\
    $(\textbf{x}_p^{n \, \times \, (p^2 \, \cdot \, c)} \leftarrow \textbf{x}_a^{h \, \times \, w \, \times \, c})$, where $n=(h \times w) / p^2$
    \STATE Transform into linear tokens and add position encoding. \\
    $\Tilde{z}_0 =  [\textbf{x}_p^{(1)}\textbf{E} ; \: \textbf{x}_p^{(2)}\textbf{E} ; \: \cdots ; \: \textbf{x}_p^{(N)}\textbf{E}] + \textbf{E}_{pos}$
    \STATE Apply self-attention to the patch embeddings to capture global features using multi-head attention with $\mathcal{K}$ heads, $\mathcal{Q}$ queries and $\mathcal{V}$ values.\\
    $S\!A = Attention (\mathcal{Q}, \, \mathcal{K}, \, \mathcal{V}) = softmax \left( \frac{\mathcal{Q}\mathcal{K}^T}{\sqrt{d_{\mathcal{K}}}} \right) \mathcal{V}$
    \STATE The final feature set is a combination of all weighted Self Attention outputs \\
    $\vec{\mathcal{F}}_{global} \leftarrow Concat(S\!A_1, S\!A_2, \dots , S\!A_H)$\textbf{W}
    \STATE\textbf{return} $\vec{\mathcal{F}}_{global}$\\
    \vspace{1mm}    
    \hspace{-2.75mm}\textbf{function:}  {SMod} $\bigr( \vec{\mathcal{F}}_{global},\,\vec{\mathcal{F}}_{spatial} \bigr)$
    \FOR {$\forall_ {1 \leq i \leq 5} \in \varphi_{S\!F\!L(i)}$ Spatial Feature Learning block} 
        \STATE Apply Max Pooling to the input $\vec{\mathcal{F}}_{global}$ featuring a kernel size of $2 \times 2$ with stride value $2$ \\
        \textit{MXP} $\leftarrow \varkappa_2^{2 \times 2}$ $[\vec{\mathcal{F}}_{global}]$
        \STATE Apply the Convolution operation with filter $f_i \in [32,64,128,256,256]$ and $3 \times 3$ kernel to Max Pooling. \\
        \textit{CONV} $\leftarrow \Psi^{3,3,f_i}(\Ddot{\mathcal{C}})$ [\textit{MXP}]
        \STATE Apply Batch Normalization after Convolution operation \\
        \textit{BN} $\leftarrow \beta$ [\textit{CONV}] 
        \STATE Apply regularization on Normalized input by randomly setting 20\% of the units to zero.\\
        \textit{DRP} $\leftarrow$ \textit{BN}
    \ENDFOR
    \STATE The output of the Spatial Feature Learning block is flattened and can be represented as: \\
    $\vec{\mathcal{F}}_{spatial} \leftarrow Flatten$ [\textit{DRP}]
    \vspace{1mm}

    \hspace{-2.75mm}\textbf{function:}  {GraFR} $\bigr( \vec{\mathcal{F}}_{spatial},\, \vec{\mathcal{F}}_{reconstructed} \bigr)$
    \STATE Nodes represent the global and spatial features.\\
    $\vec{x_v} = \left[ x_{v,1}, x_{v,2}, \cdots , x_{v,d} \right]$
    \STATE Euclidea-based edge weight computation between the feature vectors of nodes.\\
    $d_{uv}(u, v)=\sqrt{\sum_{i=1}^n \| \vec{x}_{u_i} - \vec{x}_{v_i} \|^2}$
    \STATE Hidden feature node $\hbar$ identification using the highest similarity scores. \\
    $\hbar=\{i\in V_g\:|\:\forall v \in V_g,\: (\nabla_{i \leftrightarrow v} \geq \nabla_{v \leftrightarrow i}) \}$, where $\nabla_{u \leftrightarrow v}=\left( \frac{1}{d_{uv}} \right)$
    \STATE Feature map reconstruction involves leveraging edge weights determined by similarity scores. \\
    $\mathcal{F}_{reconstructed}=\frac{\sum_{v=V_g} \, \nabla_{u \leftrightarrow v} \cdot x_v}{\sum_{v=V_g} \, \nabla_{u \leftrightarrow v}}$
    \end{algorithmic}
    
  \end{algorithm}
\end{minipage}}
\end{figure*}

\subsection{Population-based Optimization for Feature Selection}
This section explains how a population‐based meta‐heuristic algorithm is applied for feature selection and optimization. Since population-based optimization techniques randomly search for the optima of optimization problems, there is no certainty of discovering a solution in a single execution. However, as the optimization iterations and the number of random solutions increases, the likelihood of identifying the global optimum rises. Furthermore, a local search technique called Adaptive $\beta$-Hill Climbing (A$\beta$HC) is employed to refine the obtained feature subset. The optimized feature set learns a mapping between the feature set and output classes. This mapping facilitates the Hierarchical Deep Learning Classifier (HDLC) to perform the final distinction between normal cells and leukocyte cells, leveraging the optimized feature set as its input.

In this work, a population‐based meta‐heuristic algorithm, namely Sine-Cosine Algorithm (SCA)~\cite{Mirjalili2016}, is applied for feature selection and optimization. SCA follows iterative stochastic and population-based methods that imitate the harmonious behavior of sine and cosine functions to explore the solution space and find the optimal destination space. SCA typically consists of two primary stages: exploration and exploitation. The exploration phase combines the random solutions with a high degree of randomness to identify promising regions within the search space. In contrast, the exploitation phase alters the solutions with gradual modifications to reduce randomness in the variations.

To initiate the optimization procedure using SCA, the search element alters its self-position based on the sine and cosine functions as given in~\autoref{eq:scaposition}.

 \begin{equation}
\label{eq:scaposition}
   \mathcal{P}_{i, j}^{t+1} = \!
   \begin{cases}
    \!\mathcal{P}_{i, j}^{t} + r_{1,j}^{t} \times \sin{(r_{2,j}^{t})} \times |r_{3,j}^{t}D_j^t -\mathcal{P}_{i, j}^{t}|, r_{4,j}^{t} < 0.5 \\
    
    \!\mathcal{P}_{i, j}^{t} + r_{1,j}^{t} \times \cos{(r_{2,j}^{t})} \times |r_{3,j}^{t}D_j^t -\mathcal{P}_{i, j}^{t}|, r_{4,j}^{t} \geq 0.5 \end{cases}
\end{equation} 

Here, $\mathcal{P}_{i, j}^{t}$ is the position of current solution in $j^{th}$ dimension of $i^{th}$ search element at $t^{th}$ iteration. $r_{2,j}^{t}$, $r_{3,j}^{t}$ and $r_{4,j}^{t}$ are uniformly distributed random numbers, $D_j^t$ represents the position of $j^{th}$ dimension of destination point (best solution) at $t^{th}$ iteration and $||$ signify the absolute value. A random number $r_{1,j}^{t}$ facilitates the transition from exploration to exploitation of the search space, which is determined by~\autoref{eq:scarandom}.

 \begin{equation}
\label{eq:scarandom} 
    r_{1,j}^{t} = \alpha - t\frac{\alpha}{T}
\end{equation} 

Here, $\alpha$, $t$ and $T$ represents the constant number, the $t^{th}$ iteration and total number of iterations, respectively.

\begin{figure}[!ht]
  \centering
  \includegraphics[width=\columnwidth]{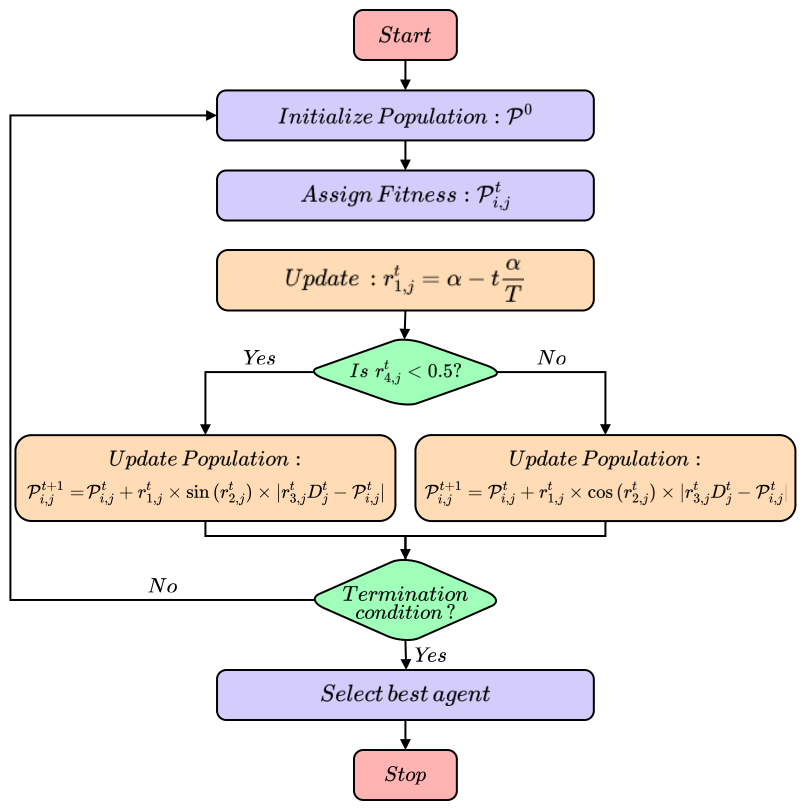}
  \caption{Flowchart of the Sine Cosine Algorithm: $\mathcal{P}_{i, j}^{t}$ denotes the $i^{th}$ element of population $\mathcal{P}$ at $t^{th}$ iteration. $\alpha$,  $T$ represents a constant number and a maximum number of iterations. $r_{1,j}^{t}$, $r_{2,j}^{t}$, $r_{3,j}^{t}$ and $r_{4,j}^{t}$ designates the random number.
  }
  \label{fig:scaflow}
\end{figure}

The value of $r_{1,j}^{t}$ decides if the search area is for exploitation (destination solution region) $(r_{1,j}^{t} \in [-1,1])$ or exploration (feasible solution region) $(r_{1,j}^{t} \in [-1,-2]$ or $r_{1,j}^{t} \in [1,2])$. Meanwhile, the stochastic variable $(r_{2,j}^{t}$, defines the search agent's movement towards or away from the destination point, bounded within $[0,2\pi]$, in sync with a complete cycle of sine and cosine functions. $(r_{3,j}^{t}$ balances the exploration and exploitation rates by bringing a random weight between $(0,2)$. Furthermore, $r_{3,j}^{t}$ introduces a stochastic step size for the destination point, either emphasizing $(r_{3,j}^{t}$>$1)$ or not emphasizing $(r_{3,j}^{t} $<$ 1)$ its impact. Finally, the parameter $r_{4,j}^{t}$ evenly transitions between the sine and cosine components, as given in~\autoref{eq:scaposition}. The flowchart of feature optimization utilizing the SCA algorithm is presented in~\autoref{fig:scaflow}.

After the identification of the most efficient features through the meta-heuristic algorithm SCA, further enhancement of exploitation ability can be achieved by integrating the local search technique named Adaptive $\beta$-Hill Climbing (A$\beta$HC). A$\beta$HC is a feature optimization algorithm utilizing local search-based techniques. These search techniques are guided by a pair of control parameters $\mathcal{N}_{\textsc{hc}}$ and $\beta_{\textsc{hc}}$, respectively. By adjusting these parameters, the search technique finds the optimal trade-off between exploitation and exploration. Fine-tuning these parameters plays a significant role in the optimization because it helps to enhance the convergence rate. The parameter $\mathcal{N}_{\textsc{hc}}$ is initially set to a value close to $1$, but it is gradually decreased as the search process iterates. This allows the algorithm to dynamically adjust $\mathcal{N}_{\textsc{hc}}$ to improve the search performance, as given in~\autoref{eq:abhcn}.

 \begin{equation}
\label{eq:abhcn} 
    \mathcal{N}_{\textsc{hc}}^t = 1 - \frac{t^{\frac{1}{P}}}{T_{max}^{\frac{1}{P}}}
\end{equation} 

Here, $\mathcal{N}_{\textsc{hc}}^t$ represents the value of $\mathcal{N}_{\textsc{hc}}$ at time $t$, $P$ is constant used to linearly decrease the value of $\mathcal{N}_{\textsc{hc}}$ to a value close to $0$ and $T_{max}$ represents the upper limit of iterations for A$\beta$HC algorithm.

Moreover, the $\beta$ parameter undergoes deterministic adaptation within a defined range $\in [\beta_{\textsc{hc}}^{min},\,\beta_{\textsc{hc}}^{max}]$, mathematically expressed in~\autoref{eq:abhcb}.

 \begin{equation}
\label{eq:abhcb} 
    \beta_{\textsc{hc}}^t = \beta_{\textsc{hc}}^{min} + t \times \frac{\beta_{\textsc{hc}}^{max} - \beta_{\textsc{hc}}^{min}}{T_{max}}
\end{equation} 

Here, $\beta_{\textsc{hc}}^t$ denotes the rate of $\beta_{\textsc{hc}}$ at time $t$, $\beta_{\textsc{hc}}^{min}$ and $\beta_{\textsc{hc}}^{max}$ represents the minimum and maximum value of $\beta_{\textsc{hc}}$ respectively, $T_{max}$ is the total number of iterations and $t$ signifies the current time. 

\subsection{Hierarchical Deep Learning Classifier}
The Hierarchical Deep Learning Classifier (HDLC) is a multi-stage DL classifier that utilizes the potential of the optimized reconstructed feature set $\mathcal{F}_{reconstructed}$. HDLC focuses on the hematological features of cells for effectively classifying both normal and leukocyte cells. HDLC is structured into two phases: an initial convolutional phase and a subsequent fully-connected phase. During the initial phase, a 1D convolutional layer employs $128$ filters with dimensions of $3 \times 3$. It computes the dot product between these filters and the input feature set denoted by $\mathcal{F}_{reconstructed}$. The convolution layer internally employs a non-linear \textit{ReLU} activation function to preserve the essential information. The output of a 1D convolution undergoes a linear transformation to produce a one-dimensional vector and subsequently input into a sequence of six fully connected dense layers in the second stage. The neurons in the fully connected layer learn non-linear associations between input features and the desired output. The final output can be interpreted as the likelihood of the input being a leukocyte cell, which is the final prediction of the classifier.

\begin{table*}[width=.9\textwidth,!ht]
\caption{Comprehensive Dataset Description: The experimental configuration includes four distinct datasets, each correlated with unique cell types.
% Dataset Overview. The experimental setup incorporates four distinct datasets, each associated with unique cell types.
}
    \begin{tabular*}{\tblwidth}{@{} LCCCCCC@{} }
    \toprule
        \textbf{{Dataset}}
        &\multicolumn{2}{c}{\textbf{{Original Leukocyte}}} 
        &\multicolumn{2}{c}{\textbf{{Synthetic Leukocyte}}} 
        &\multicolumn{2}{c}{\textbf{{Total Leukocyte}}} \\ 
        
        \cmidrule(l){2-3} \cmidrule(l){4-5} \cmidrule(l){6-7}
            &\textbf{Cancer Class}&\textbf{Normal Class}
            &\textbf{Cancer Class}&\textbf{Normal Class}
            &\textbf{Cancer Class}&\textbf{Normal Class}\\         
    \midrule
        C-NMC 2019 & 8491 & 4037 & NIL & 4454 & 8491 & 8491  \\
        ALL-IDB1   & 49   & 59   & NIL & NIL    & 49   & 59  \\
        ALL-IDB2   & 130  & 130  & NIL & NIL    & 130  & 130  \\
        SN-AM      & 30   & 30   & NIL & NIL    & 30   & 30  \\
    \bottomrule    
    \multicolumn{7}{@{}l}{\footnotesize `NIL' represents the absence of synthetic leukocyte generation.}    
    \end{tabular*}
    \\
    \label{table:dataset-details}  
\end{table*}

\section{Experimental Evaluations}
\label{sec:exp-eval}
This section illustrates the diverse datasets used for experimental evaluations of the proposed CoTCoNet framework for classifying leukemic cells on blood cell images. Subsequently, we present the experimental setup, which includes the comparison methods, along with the evaluation metrics used. Finally, experimental results show the effectiveness of our framework. 

\subsection{Dataset Summarization}
\label{sec:datasetsummary}
The proposed CoTCoNet framework was tested using four publicly available datasets, namely, C-NMC 2019~\cite{Gupta2022}, ALL-IDB1~\cite{Labati2011}, ALL-IDB2~\cite{Labati2011}, and SN-AM datasets~\cite{Gupta2020}. The leukocyte samples used for experimentation are tabulated in \autoref{table:dataset-details}. We briefly summarize the dataset as follows:

% \noindent
\textbf{C-NMC 2019:} It is also known as the ALL Challenge dataset of ISBI 2019, a collection of microscopic images of bone marrow from healthy participants and participants with Acute Lymphoblastic Leukemia (ALL). C-NMC 2019 was created to broaden the scope of research on differentiating leukemic blasts from normal cells, as both appear similar under a microscope. The leukemic cells in the dataset are labeled as ``all'', and normal cells are labeled as ``hem''.

% \noindent
\textbf{ALL-IDB1:} Acute Lymphoblastic Leukemia Image Database (ALL-IDB1) for Image Processing initiative resulted in datasets for segmentation and image classification. It consists of multi-cell blood microscopic images and is divided into two main categories: ALL-positive (blood samples with leukemic cells), labeled as ``ImXXX\_1.jpg'' and ALL-negative (blood samples from non-ALL patients), labeled as ``ImXXX\_0.jpg''. Expert oncologists have labeled the lymphocytes in these blood samples.

% \noindent
\textbf{ALL-IDB2:} The images in this dataset are cropped areas of interest of normal and leukemic cells from ALL-IDB1. It is a single-cell blood microscopic image for evaluating leukemic cell classification systems.

% \noindent
\textbf{SN-AM:} The microscopic images in this dataset are collected from patients using a bone marrow aspiration procedure. The bone marrow slides of patients diagnosed with B-lineage Acute Lymphoblastic Leukemia (B-ALL) and Multiple Myeloma (MM).

% \begin{table*}[width=.9\textwidth,!ht]
% \caption{Dataset Sample Details}
% \begin{tabular*}{\tblwidth}{@{} LCCCCCC@{} }
% \toprule
% \textbf{Dataset} & \multicolumn{2}{c}{\textbf{Train}} & \multicolumn{2}{c}{\textbf{Test}} & \multicolumn{2}{c}{\textbf{Validation}} \\ 
% \cmidrule(l){2-3} \cmidrule(l){4-5} \cmidrule(l){6-7}
% &\textbf{Normal}&\textbf{Affected}
% &\textbf{Normal}&\textbf{Affected}            
% &\textbf{Normal}& \textbf{Affected}\\
% % &&\textbf{Tuberculosis*}&&\textbf{Tuberculosis }&&\textbf{Tuberculosis}\\
% \midrule
% C-NMC Dataset &3410 &3419 &398 &427 &399 &427\\
% ALL-IDB1 Dataset &3480 &3420 &440 &440 &440 &440\\
% ALL-IDB1 Dataset    &2800 &2864 &350 &358 &350 &358\\
% SNAM Dataset    &2800 &2864 &350 &358 &350 &358\\
% \bottomrule    
% \multicolumn{7}{@{}l}{$^\#$\footnotesize Pneumonia, $^*$\footnotesize Tuberculosis}    
% \end{tabular*}
% \\  
% \label{table:dataset-details}    
% \end{table*}

\begin{table*}[width=.9\textwidth,!b]
\caption{
Performance Comparison on C-NMC 2019 Dataset.}
    \begin{tabular*}{\tblwidth}{@{} LLCCCC@{} }
        \toprule
        \multicolumn{1}{c}{} & \textbf{ Methods} & \textbf{Acc$^a$} & \textbf{ Pre$^b$} & \textbf{ Rec$^c$} & \textbf{ F1$^d$} \\
        \midrule
        \multicolumn{1}{l}{\multirow{4}{*}{Deep Learning-based Classification}} & Mohammed~\textit{et~al.}~\cite{Mohammed2023} & 0.9629 & 0.9793 & 0.9458 & 0.9623 \\
        \multicolumn{1}{c}{} & Gehlot~\textit{et~al.}~\cite{Gehlot2020} & 0.9630 & 0.9554 & 0.9291 & 0.9480 \\
        \multicolumn{1}{c}{} & Jawahar~\textit{et~al.}~\cite{Jawahar2022} & 0.9100 & 0.9300 & 0.9800 & 0.9600 \\
        \multicolumn{1}{c}{} & Das~\textit{et~al.}~\cite{Das2022} & 0.9529 & 0.9282 & 0.9765 & 0.9517 \\
        \midrule
        \multicolumn{1}{l}{\multirow{1}{*}{Imbalanced Classification}} & Depto~\textit{et~al.}~\cite{Depto2023} & 0.7729 & 0.7803 & 0.7729 & 0.7766 \\
        \midrule
        \multicolumn{1}{l}{\multirow{1}{*}{Segmentation-based Classification}} & Khandekar~\textit{et~al.}~\cite{Khandekar2021} & 0.9710 & 0.9720 & 0.9701 & 0.9710 \\
        \midrule
        \multicolumn{1}{c}{} & \textbf{CoTCoNet} & \textbf{0.9894} & \textbf{0.9800} & \textbf{0.9988} & \textbf{0.9893} \\
        \bottomrule
        \multicolumn{6}{l}{\scriptsize $^a$ Accuracy, $^b$ Precision, $^c$ Recall, $^d$ F1-Score}
    \end{tabular*}
\label{table:sota-cnmc}
\end{table*}

\subsection{Experimental Setup}
\label{sec:exp setup}

In this section, we present a concise overview of the current state-of-the-art methods used as benchmarks to assess the effectiveness of our proposed methods. We also discuss the evaluation metrics used for appraising the performance of different methods.

\subsubsection{Comparison Methods}
\label{sec:comparison-methods}
We compare our framework against three different types of methods.

% \paragraph{\textbf{Deep Learning-based Classification and Detection Methods:} }
\noindent \textbf{i) Deep Learning-based Classification :} The existing approach for Acute Lymphoblastic Leukemia (ALL) classification includes Transfer Learning and Convolutional Neural Networks (CNN) architecture as their backbone. Mohammed~\textit{et~al.}~\cite{Mohammed2023} propose an ensemble strategy that combines a CNN with Bidirectional Long Short-Term Memory (BiLSTM) and a Gated Recurrent Unit (GRU) architecture in conjunction with Multiclass Support Vector Machine (MSVM) classifier. Gehlot~\textit{et~al.}~\cite{Gehlot2020} incorporate a stain deconvolutional CNN alongside a Kernel SVM auxiliary classifier to harness spectral-averaged features. Jawahar~\textit{et~al.}~\cite{Jawahar2022} employs depth-wise convolution with diverse dilation rates to extract robust global and local features from microscopic bone marrow slides. Das~\textit{et~al.}~\cite{Das2022} use ResNet18-based architecture supplemented with an orthogonal softmax layer for ALL detection.

% \begin{enumerate}[i)]
%     \item \textbf{Deep Learning-based Classification :}
% \end{enumerate}

% \paragraph{Imbalanced Classification Methods for Leukemia Detection: }
\noindent \textbf{ii) Imbalanced Classification:} This includes comprehensive qualitative and quantitative assessments of imbalanced leukemia datasets. Depto~\textit{et~al.}~\cite{Depto2023} handles class imbalance issues by incorporating Generative Adversarial Networks (GANs) and integrating DL methods to introduce a bias towards the majority class. Their approach adopts DenseNet121 primary architecture and integrates a class-weighted loss technique for both the original hematological cell images and conditional GAN-generated images.

% \paragraph{Image Analysis and Object Detection for Leukemia Diagnosis: }
\noindent \textbf{iii) Segmentation-based Classification:} This involves segmentation techniques to predict leukemic cells within microscopic blood smear images. Khandekar~\textit{et~al.}~\cite{Khandekar2021} employ You Only Look Once (YOLOv4) in combination with CSPDarkNet53 for feature extraction. Additionally, the author incorporates Spatial Pyramid Pooling and the Path Aggregation Network to enhance feature extraction and produce the output along with its corresponding class label. 

\subsubsection{Evaluation Metrics}
\label{sec:evaluation-metrics}
To assess the performance of our framework, we adopt widely used evaluation metrics for leukemic cell classification, such as accuracy, precision, recall, and $F1$-Score. We focus on accuracy and $F1$-Score because our dataset is not skewed towards the majority or minority class. We also consider precision and recall to visualize the model's ability to make accurate positive predictions and identify all relevant positive instances. \autoref{table:evaluation-metrics} shows the mathematical representations of the metrics.

% \begin{table}[width=\linewidth,cols=2,pos=b]
\begin{table}[width=0.9\columnwidth, pos=ht]
    \caption{Evaluation metrics for CoTCoNet framework in leukocyte and normal cell classification.   
    % Measures for assessing CoTCoNet framework for leukocyte cell and normal cell classification.
    % Measures for binary classification of leukocyte cell and normal cell.
    % , such that $T\!P_{leuk}$ - True Positive, $T\!N_{nor}$ - True Negative, $F\!P_{leuk}$ - False Positive, and $F\!N_{nor}$ - False Negative respectively.
    }
    \begin{tabular*}{\tblwidth}{@{} LC@{} }
        \toprule
        \textbf{Evaluation Metrics} & \textbf{Mathematical Formulation} \\
        \midrule
        Accuracy & $ \mathlarger{\frac{T\!P_{leuk}+T\!N_{nor}}{T\!P_{leuk}+F\!P_{leuk}+T\!N_{nor}+F\!N_{nor}}}$  \\
        & \\
        Precision & $ \mathlarger{\frac{T\!P_{leuk}}{T\!P_{leuk} +F\!P_{leuk}}}$  \\
        & \\
        Recall & $ \mathlarger{\frac{T\!P_{leuk}}{T\!P_{leuk}+F\!N_{nor}}}$  \\
        & \\
        F1 - Score & $ \mathlarger{ \frac{2 \times Precision \times Recall}{Precision + Recall}}$ \\
        \bottomrule
    \end{tabular*}
\label{table:evaluation-metrics} 
\end{table}

% \begin{align*} 
% Accuracy &= \frac{T\!P_{leuk}+T\!N_{nor}}{T\!P_{leuk}+F\!P_{leuk}+T\!N_{nor}+F\!N_{nor}}\\ 
% Precision &= \frac{T\!P_{leuk}}{T\!P_{leuk} +F\!P_{leuk}} \\
% Recall &= \frac{T\!P_{leuk}}{T\!P_{leuk}+F\!N_{nor}} \\
% F1 - Score &= 2 \times \frac{Precision \times Recall}{Precision + Recall}
% \end{align*}

Here, $T\!P_{leuk}$ is True Positive, a leukemia sample correctly predicted as leukemia. $T\!N_{nor}$ is True Negative, a non-leukemia sample correctly predicted as non-leukemia. $F\!P_{leuk}$ is False Positive, a non-leukemia sample incorrectly predicted as leukemia. $F\!N_{nor}$ is False Negative, a leukemia sample incorrectly predicted as non-leukemia.

% Accuracy is the proportion of all correct predictions. Precision is the proportion of all predicted leukemia samples that are true leukemia samples. Recall is the proportion of all leukemia samples correctly predicted as leukemia. $F1$-Score is the harmonic mean of precision and recall, which balances both metrics equally.

\subsection{Experimental Results}
\label{sec:experimental-results}
This section presents the experimental results and comparisons of our proposed method. First, we evaluate the performance of our method against state-of-the-art methods. Next, we perform ablation studies to evaluate the impact and effectiveness of individual modules consisting of our framework.

hl\subsubsection{Performance Evaluation}
\label{sec:performance-evaluation}
We compare the proposed CoTCoNet framework with the existing methods on different datasets to show its performance.

\vspace{1.5mm}
\noindent \textbf{Effective Comparison on C-NMC 2019 Dataset:}
\autoref{table:sota-cnmc} presents a detailed overview of the state-of-the-art methods in leukemic cell classification and shows how our proposed method outperforms other methods.

We can observe that Deep Learning-based Classification (proposed by Mohammed~\textit{et~al.}~\cite{Mohammed2023}, Gehlot~\textit{et~al.}~\cite{Gehlot2020}, Jawahar~\textit{et~al.}~\cite{Jawahar2022} and Das~\textit{et~al.}~\cite{Das2022}) does not yield favorable performance in comparison to CoTCoNet. Particularly, Jawahar~\textit{et~al.}~\cite{Jawahar2022} approach exhibits a lower performance, with a decrease of 7.94\% due to the limited effectiveness of its cluster layers in capturing both local and global features efficiently. On the other hand,~\cite{Das2022} struggles to capture the hidden discriminative features within leukocyte cell images, resulting in a 3.65\% decrease in accuracy. The methods proposed by~\cite{Mohammed2023} and~\cite{Gehlot2020} both combine a feature extraction module with a Support Vector Machine (SVM) classifier, which results in satisfactory performance but not as good as CoTCoNet. However, CoTCoNet integrates spatial and global features and identifies the hidden features to achieve an overall improvement of 2-7\% in accuracy and 2-4\% in F1-Score compared to these methods. Depto~\textit{et~al.}~\cite{Das2022} method offers potential for improvement compared to other methods, primarily due to its handling of imbalanced data samples. Conversely, Khandekar~\textit{et~al.}~\cite{Khandekar2021} demonstrates competitive performance by incorporating segmentation techniques with feature extraction methods,  resulting in a minor accuracy drop of 1.84\%. Our proposed CoTCoNet framework addresses the class imbalance issue by employing a GAN-driven mechanism for cell synthesis and segmenting leukocyte cells while excluding extraneous sections, such as plasma, to enhance feature extraction.

\begin{table*}[width=.9\textwidth,pos=ht]
\caption{Performance Comparison on ALL-IDB1 Dataset.}
    \begin{tabular*}{\tblwidth}{@{} LLCCCC@{} }
      \toprule
      \multicolumn{1}{c}{} & \textbf{ Methods} & \textbf{Acc$^a$} & \textbf{ Pre$^b$} & \textbf{ Rec$^c$} & \textbf{ F1$^d$} \\
      \midrule
      \multicolumn{1}{l}{\multirow{4}{*}{Deep Learning-based Classification}} & Mohammed~\textit{et~al.}~\cite{Mohammed2023} & 0.9412 & 0.8889 & 1.0000 & 0.9412 \\
      \multicolumn{1}{c}{} & \textbf{Gehlot~\textit{et~al.}}~\cite{Gehlot2020} & \textbf{1.0000} & \textbf{1.0000} & \textbf{1.0000} & \textbf{1.0000} \\
      \multicolumn{1}{c}{} & Jawahar~\textit{et~al.}~\cite{Jawahar2022} & 0.9412 & 0.8750 & 1.0000 & 0.9333 \\
      \multicolumn{1}{c}{} & Das~\textit{et~al.}~\cite{Das2022} & 0.9939 & 1.0000 & 0.9865 & 0.9932 \\
      \midrule
      \multicolumn{1}{l}{\multirow{1}{*}{Imbalanced Classification}} & Depto~\textit{et~al.}~\cite{Depto2023} & 0.9118 & 0.9444 & 0.8947 & 0.9189 \\
      \midrule
      \multicolumn{1}{l}{\multirow{1}{*}{Segmentation-based Classification}} & Khandekar~\textit{et~al.}~\cite{Khandekar2021} & 0.9787 & 0.9844 & 0.9844 & 0.9844 \\
      \midrule
      \multicolumn{1}{c}{} & \textbf{CoTCoNet} & \textbf{1.0000} & \textbf{1.0000} & \textbf{1.0000} & \textbf{1.0000} \\
      \bottomrule
      \multicolumn{6}{l}{\scriptsize $^a$ Accuracy, $^b$ Precision, $^c$ Recall, $^d$ F1-Score}
    \end{tabular*}
\label{table:sota-all1}
\end{table*}

\begin{table*}[width=.9\textwidth,!b]
\caption{Performance Comparison on ALL-IDB2 Dataset.}
    \begin{tabular*}{\tblwidth}{@{} LLCCCC@{} }
      \toprule
      \multicolumn{1}{c}{} & \textbf{ Methods} & \textbf{Acc$^a$} & \textbf{Pre$^b$} & \textbf{ Rec$^c$} & \textbf{ F1$^d$} \\
      \midrule
      \multicolumn{1}{l}{\multirow{4}{*}{Deep Learning-based Classification}} & Mohammed~\textit{et~al.}~\cite{Mohammed2023} & 0.9750 & 1.0000 & 0.9524 & 0.9756 \\
      \multicolumn{1}{c}{} & Gehlot~\textit{et~al.}~\cite{Gehlot2020} & 0.9250 & 0.9500 & 0.9048 & 0.9268 \\
      \multicolumn{1}{c}{} & Jawahar~\textit{et~al.}~\cite{Jawahar2022} & 0.8250 & 0.9500 & 0.76.00 & 0.8444 \\
      \multicolumn{1}{c}{} & Das~\textit{et~al.}~\cite{Das2022} & 0.9821 & 0.9845 & 0.9795 & 0.9820 \\
      \midrule
      \multicolumn{1}{l}{\multirow{1}{*}{Imbalanced Classification}} & Depto~\textit{et~al.}~\cite{Depto2023} & 0.8167 & 0.8659 & 0.8417 & 0.8536 \\
      \midrule
      \multicolumn{1}{l}{\multirow{1}{*}{Segmentation-based Classification}} & Khandekar~\textit{et~al.}~\cite{Khandekar2021} & 0.9787 & 0.9844 & 0.9844 & 0.9844 \\
      \midrule
      \multicolumn{1}{c}{} & \textbf{CoTCoNet} & \textbf{1.0000} & \textbf{1.0000} & \textbf{1.0000} & \textbf{1.0000} \\
      \bottomrule
      \multicolumn{6}{l}{\scriptsize $^a$ Accuracy, $^b$ Precision, $^c$ Recall, $^d$ F1-Score}
    \end{tabular*}
  \label{table:sota-all2}
\end{table*}

\vspace{1.5mm}
\noindent \textbf{Effective Comparison on ALL-IDB1 Dataset:}
The performance evaluation of the CoTCoNet framework relative to existing methods on the ALL-IDB1 dataset can be observed in~\autoref{table:sota-all1}. The method proposed by Depto~\textit{et~al.}~\cite{Depto2023} addresses the imbalance in the ALL-IDB1 dataset, resulting in a noticeable reduction in accuracy of 8.82\%. Khandekar~\textit{et~al.}~\cite{Khandekar2021} proposes a method that classifies cells based on regional features, demonstrating commendable performance with a marginal accuracy variation of merely 2.13\%. CoTCoNet, in contrast, uses a balanced dataset and gives equal importance to geometrical and global features, leading to improved classification performance. Among Deep Learning-based Classification, Gehlot~\textit{et~al.}~\cite{Gehlot2020} and CoTCoNet framework outperforms other methods due to their focus on feature-based deep classifier. However, the methods by Mohammed~\textit{et~al.}~\cite{Mohammed2023} and Jawahar~\textit{et~al.}~\cite{Jawahar2022} do not efficiently utilize the prime features of cell images, which results in accuracy gap of 5.88\%. Meanwhile, Das~\textit{et~al.}~\cite{Das2022} achieve comparable performance, with a mere 0.61\% difference in accuracy, when utilizing a residual-based architecture for feature learning. However, when compared to these techniques, the CoTCoNet framework is curated to learn the hidden features of cell images for precise classification, which gives it an edge over the other techniques.

\vspace{1.5mm}
\noindent \textbf{Effective Comparison on ALL-IDB2 Dataset:}

Comparison results on the ALL-IDB2 dataset are presented in~\autoref{table:sota-all2}. The study conducted by Depto~\textit{et~al.}~\cite{Depto2023} discusses an imbalanced scenario in leukemia classification, where their method learns fewer features, resulting in a significant decrease in performance. Khandekar~\textit{et~al.}~\cite{Khandekar2021} effectively segments overlapped cells in clustered regions, leading to satisfactory classification performance. CoTCoNet, on the other hand, takes a comprehensive approach by incorporating GAN-driven solutions to tackle class imbalance problems and segmenting blood smear images to learn sharp features for precise classification. The Deep Learning classification method that Jawahar~\textit{et~al.}~\cite{Jawahar2022} utilizes does not yield promising results, this method fails to emphasize prominent hidden features. Furthermore, techniques by Mohammed~\textit{et~al.}~\cite{Mohammed2023} and Das~\textit{et~al.}~\cite{Das2022} exhibit almost identical performance, as the author uses convolution-based filters to obtain features but fail to combine spatial and global features which leads to marginal accuracy decrease in the range of 1-3\%. The method proposed by Gehlot~\textit{et~al.}~\cite{Gehlot2020} utilizes spectral-averaged features constrainedly, resulting in a 7.5\% performance decline. However, CoTCoNet demonstrates superior performance by optimizing the learning of features, thereby significantly improving the overall performance for the classification of leukocytes.

\begin{figure*}[pos=bp]
  \centering
  \includegraphics[width=\textwidth]{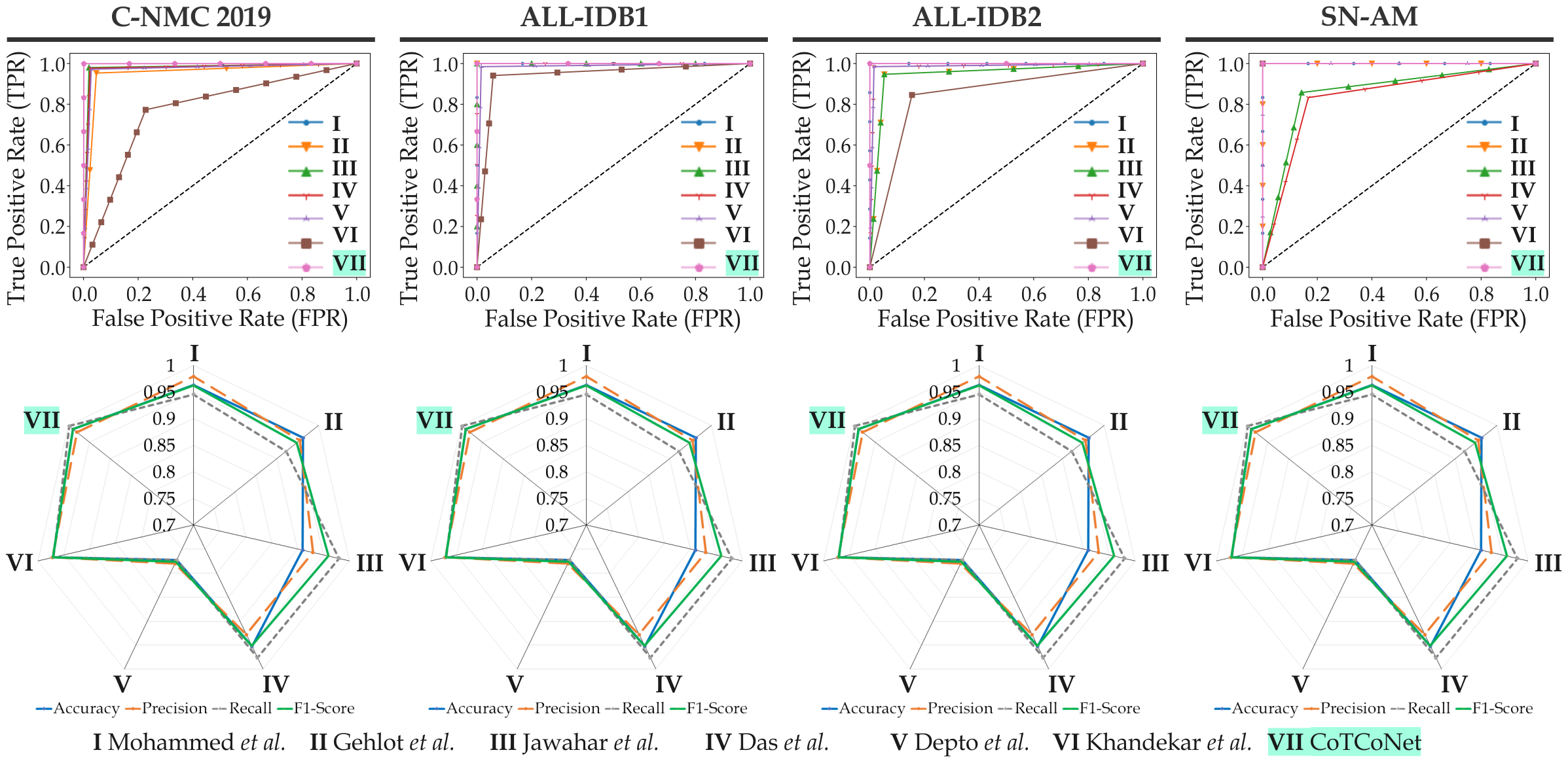}
  \caption{Comparative Analysis of CoTCoNet framework and existing methods across four datasets. The ROC (Receiver Operating Characteristic) is used to evaluate performance. Radar charts depict the performance of each method across accuracy, precision, recall, and F1-Score metrics across various datasets.}
  \label{fig:performance-comparison}
\end{figure*}

% \begin{table*}[width=.9\textwidth,pos=h]
\begin{table*}[width=.9\textwidth,pos=bp]
\caption{Performance Comparison on SN-AM Dataset.}
    \begin{tabular*}{\tblwidth}{@{} LLCCCC@{} }
      \toprule
      \multicolumn{1}{c}{} & \textbf{ Methods} & \textbf{Acc$^a$} & \textbf{ Pre$^b$} & \textbf{ Rec$^c$} & \textbf{ F1$^d$} \\
      \midrule
      \multicolumn{1}{l}{\multirow{4}{*}{Deep Learning-based Classification}} & Mohammed~\textit{et~al.}~\cite{Mohammed2023} & 0.8462 & 1.0000 & 0.7778 & 0.8750 \\
      \multicolumn{1}{c}{} & Gehlot~\textit{et~al.}~\cite{Gehlot2020} & 0.9231 & 0.8571 & 1.0000 & 0.9231 \\
      \multicolumn{1}{c}{} & Jawahar~\textit{et~al.}~\cite{Jawahar2022} & 0.8462 & 0.8571 & 0.8571 & 0.8571 \\
      \multicolumn{1}{c}{} & Das~\textit{et~al.}~\cite{Das2022} & 0.7692 & 0.7143 & 0.8333 & 0.7692 \\
      \midrule
      \multicolumn{1}{l}{\multirow{1}{*}{Imbalanced Classification}} & Depto~\textit{et~al.}~\cite{Depto2023} & 0.8769 & 0.8142 & 0.9521 & 0.8778 \\
      \midrule
      \multicolumn{1}{l}{\multirow{1}{*}{Segmentation-based Classification}} & Khandekar~\textit{et~al.}~\cite{Khandekar2021} & 0.9231 & 1.0000 & 0.8750 & 0.9333 \\
      \midrule      
      \multicolumn{1}{c}{} & \textbf{CoTCoNet} & \textbf{1.0000} & \textbf{1.000} & \textbf{1.0000} & \textbf{1.0000} \\
      \bottomrule
      \multicolumn{6}{l}{\scriptsize $^a$ Accuracy, $^b$ Precision, $^c$ Recall, $^d$ F1-Score}
    \end{tabular*}
  \label{table:sota-snam}
\end{table*}

\vspace{1.5mm}
\noindent \textbf{Effective Comparison on SN-AM Dataset:}
The performance of the CoTCoNet framework and other state-of-the-art techniques in leukemic cell classification is presented in~\autoref{table:sota-snam}. Several studies, including Das~\textit{et~al.}~\cite{Das2022}, Mohammed \textit{et al.} \cite{Mohammed2023} and Jawahar~\textit{et~al.}~\cite{Jawahar2022} overlook feature selection and optimization, resulting in lower performance. In contrast, Gehlot~\textit{et~al.}~\cite{Gehlot2020} effectively harnesses spectral-averaged features of blood cells to achieve superior performance among Deep Learning-based methods. Khandekar~\textit{et~al.}~\cite{Khandekar2021} employs a region-based segmentation technique to classify leukemic cells. On the other hand, Depto~\textit{et~al.}~\cite{Depto2023} demonstrate that the classification performance is degraded with imbalanced data samples. However, CoTCONet leverages a GAN-driven technique to decode class imbalance issues effectively. Additionally, a segmentation technique is employed to extract leukocyte cells to learn a  robust mapping between features and classes. This mapping enables the feature-learning mechanism to quantify the core features for better discrimination of the leukemic class and non-leukemic class. As a result, our proposed framework exhibits superior performance when applied to the SN-AM dataset compared to existing techniques. \autoref{fig:performance-comparison} shows the Receiver Operating Characteristic (ROC) curve and the performance comparison of different methods.

\begin{table*}[width=.9\textwidth,pos=h]
\caption{Ablation Study on CNMC Dataset.}
    \begin{tabular*}{\tblwidth}{@{} LCCCC@{} }
    % \toprule
        \textbf{ Methods} & \textbf{Acc$^a$} & \textbf{ Pre$^b$} & \textbf{ Rec$^c$} & \textbf{ F1$^d$} \\
        \midrule
        % CLAHE + SHARPEN
        \textbf{Effect of Image Enhancement} & & & & \\
        \midrule
        CoTCoNet (No Enhancement) & 0.9011 & 0.8681 & 0.9294 & 0.8977 \\
        CoTCoNet (Fuzzy)          & 0.9211 & 0.9741 & 0.8807 & 0.9251 \\
        CoTCoNet (MirNET)         & 0.9499 & 0.8999 & 1.0000 & 0.9473 \\
        CoTCoNet (HE)             & 0.9541 & 0.9753 & 0.9356 & 0.9550 \\
        CoTCoNet (Sharpen)        & 0.9611 & 0.9435 & 0.9780 & 0.9604 \\
        CoTCoNet (CLAHE)          & 0.9788 & 0.9647 & 0.9927 & 0.9785 \\
        % & & & & \\

        \midrule
        \textbf{Effect of Segmentatiom} & & & & \\
        \midrule
        CoTCoNet (UNet)   & 0.8657 & 0.8092 & 0.9124 & 0.8577 \\
        CoTCoNet (UNet++) & 0.8920 & 0.6568 & 0.9839 & 0.7878 \\
        % & & & & \\

        \midrule
        \textbf{Effect of Augmentation} & & & & \\
        \midrule
        CoTCoNet (No Augmentation) & 0.8206 & 0.5457 & 0.8024 & 0.6496 \\
        CoTCoNet (Traditional Augmentation) & 0.8767 & 0.8499 & 0.8919 & 0.8704 \\
        % & & & & \\

        \midrule
        \textbf{Effect of Feature Extraction} & & & & \\
        \midrule
        CoTCoNet (ViT-B/32)             & 0.8581 & 0.8269 & 0.8819 & 0.8535 \\
        CoTCoNet (ViT-L/32)             & 0.8422 & 0.8996 & 0.7703 & 0.8299 \\
        CoTCoNet (ViT-H/16)             & 0.8351 & 0.6914 & 0.9702 & 0.8075 \\
        CoTCoNet (GMod)              & 0.8651 & 0.8422 & 0.8827 & 0.862  \\
        CoTCoNet (VGG-16)            & 0.8528 & 0.8492 & 0.8303 & 0.8396 \\
        CoTCoNet (SMod)              & 0.8687 & 0.9184 & 0.8092 & 0.8604 \\
        % CoTCoNet (ResNet-101)           & 0.8928 & 0.8492 & 0.9303 & 0.8879 \\
        CoTCoNet (GMod + SMod) & 0.9395 & 0.8898 & 0.9883 & 0.9365 \\
        % & & & & \\

        \midrule        
        \textbf{Effect of Graph-based Feature Reconstruction and Feature Selection} & & & & \\
        \midrule
        CoTCoNet (w\textbackslash o graph + w\textbackslash o selection) & 0.9435 & 0.9764 & 0.9160 & 0.9453 \\
        CoTCoNet (with graph + w\textbackslash o selection)           & 0.9510 & 0.9494 & 0.9988 & 0.9734 \\
        CoTCoNet (w\textbackslash o graph + with selection)           & 0.9865 & 0.9741 & 0.9988 & 0.9863 \\
        
        \midrule        
        \textbf{CoTCoNet} & \textbf{0.9894} & \textbf{0.9800} & \textbf{0.9988} & \textbf{0.9893} \\
        \bottomrule
        \multicolumn{5}{l}{\scriptsize $^a$ Accuracy, $^b$ Precision, $^c$ Recall, $^d$ F1-Score, w/o shorthand for ``without''}
    \end{tabular*}
\label{table:ablation}
\end{table*}

\subsubsection{Ablation Study}
\label{sec:ablation-study}
% \S Revisiting Computer-Aided Tuberculosis Diagnosis PAMI
To assess the contributions of the key modules within the CoTCoNet framework, we systematically removed each module one at a time on the C-NMC 2019 Dataset (refer~\autoref{table:ablation}). This ablation study showed that all proposed components and strategies actively enhance the overall performance of our framework. In~\autoref{table:ablation}, the additional technique or module under examination is indicated within parentheses (e.g., CoTCoNet [technique/module]). Specifically, we explored the effects of different components, such as Image Enhancement, Segmentation, Data Augmentation, Feature Extraction, Graph-based Feature Reconstruction, and Feature Selection. The pipeline for our proposed model is comprehensively described in \autoref{sec:preprocessing}.

\textit{Effect of Image Enhancement:} The ablation study in \autoref{table:ablation} shows that the accuracy of CoTCoNet drops by 8.83\% when no image enhancement technique is applied. Other enhancement techniques, such as Fuzzy~\cite{Kadak2023}, MirNet~\cite{Zamir2023}, and Histogram Equalization~\cite{Rahman2023}, yield accuracy reductions ranging from 3-6\%. CoTCoNet leverages the combined use of CLAHE, followed by the Sharpen technique. The performance decreases by 2.83\% without Sharpen and by 1.06\% without CLAHE. These results show a strong dependency on image enhancement techniques within the CoTCoNet framework or enhancing the learning of features from leukocytes.

\textit{Effect of Segmentation:} \autoref{table:ablation} shows that CoTCoNet achieves a significantly higher accuracy of 12.57\% and 9.27\% than CoTCoNet(UNet~\cite{Ronneberger2015}) and CoTCoNet (UNet++ \cite{Zhou2018}), respectively. CoTCoNet outperforms the above techniques and takes advantage of the finely-tuned Leukemia Segment Anything Model (LeuSAM) to precisely identify leukocyte cell linings. This provides a good initialization point for parameter learning, contributing to the superior performance of CoTCoNet.

\textit{Effect of Augmentation:} Examining the data augmentation in~\autoref{table:ablation}, it is apparent that utilizing imbalanced leukemia samples leads to a substantial accuracy drop of 16.88\%, primarily due to overfitting. Conversely, employing CoTCoNet (Traditional Augmentation), which includes scaling, cropping, translation, flipping, and rotation, there is a notable performance improvement of 5.61\%. However, this improvement is 11.27\% lower than what is achieved by our proposed CoTCoNet framework. These findings demonstrate the positive impact of data augmentation techniques to enhance method performance. Unlike the previous techniques, CoTCoNet incorporates LeuGAN with the aim of synthesizing high-quality leukocyte cells to balance the data samples, making the model more robust and trainable.

\textit{Effect of Feature Extraction:} To assess the importance of the feature extractor, we perform an ablation experiment using only transformer-based and convolution-based feature extractors (i.e., without Feature Reconstruction and Selection). The results in~\autoref{table:ablation} indicate that performance declines by 13-15\% in transformer-based feature extractors, specifically ViT-B/32, ViT-L/32, and ViT-H/16. Interestingly, the utilization of a custom feature extractor known as the Global Module (GMod) demonstrates a 3-7\% improvement over transformer-based extractors. Furthermore, the accuracy of the convolution-based VGG-16 feature module is 1.59\% less accurate compared to the custom convolution-based feature extractor, Spatial Module (SMod). However, the coupled performance of both GMod and SMod surpasses that of the other feature extractors because they learn from the integrated global and spatial features of the leukocyte cells.

\textit{Effect of Graph-based Feature Reconstruction and Feature Selection:} 
\autoref{table:ablation} demonstrates the supremacy of the CoTCoNet framework, which employs a graph-based mechanism for feature reconstruction and a population-based algorithm for feature selection and optimization. Specifically, CoTCoNet (w\textbackslash o graph + with selection) achieves competitive performance, outperforming the proposed CoTCoNet by a margin of 0.29\%. This observation highlights the significance of feature selection in the training process. Conversely, alternative combinations (i.e., w\textbackslash o graph + w\textbackslash o selection and with graph + w\textbackslash o selection) failed to deliver effective performance on classifying leukemic cells.

\begin{figure*}[!ht]
    \centering
    \begin{minipage}[b]{0.45\linewidth}% Figure 1
        \includegraphics[width=\textwidth]{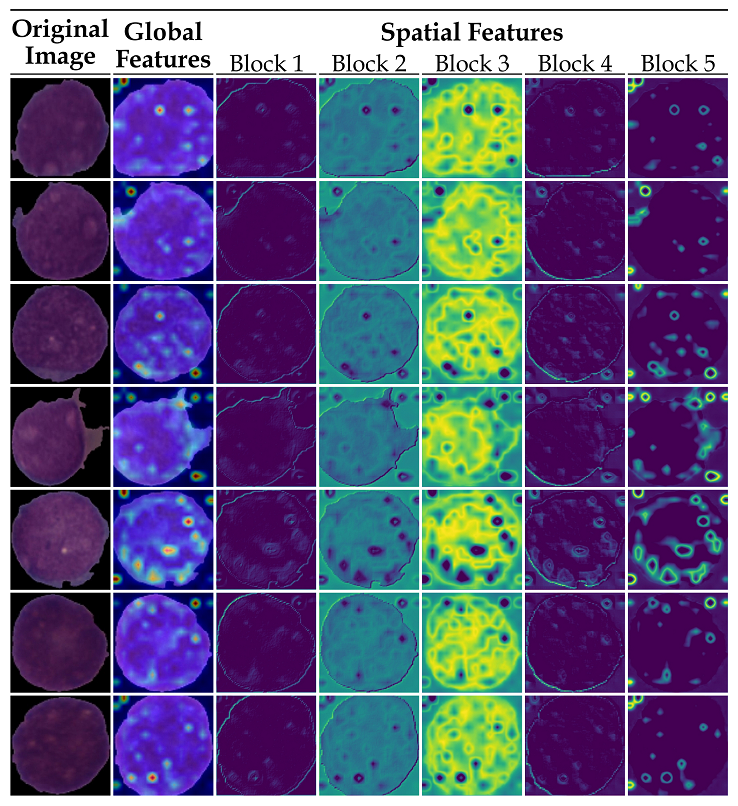}
        %\caption{default}
        \centering
        \small (a)
        %\label{fig:figure1}
    \end{minipage}
    %\hspace{0.5cm}
    \begin{minipage}[b]{0.45\linewidth}% Figure 2
        \includegraphics[width=\textwidth]{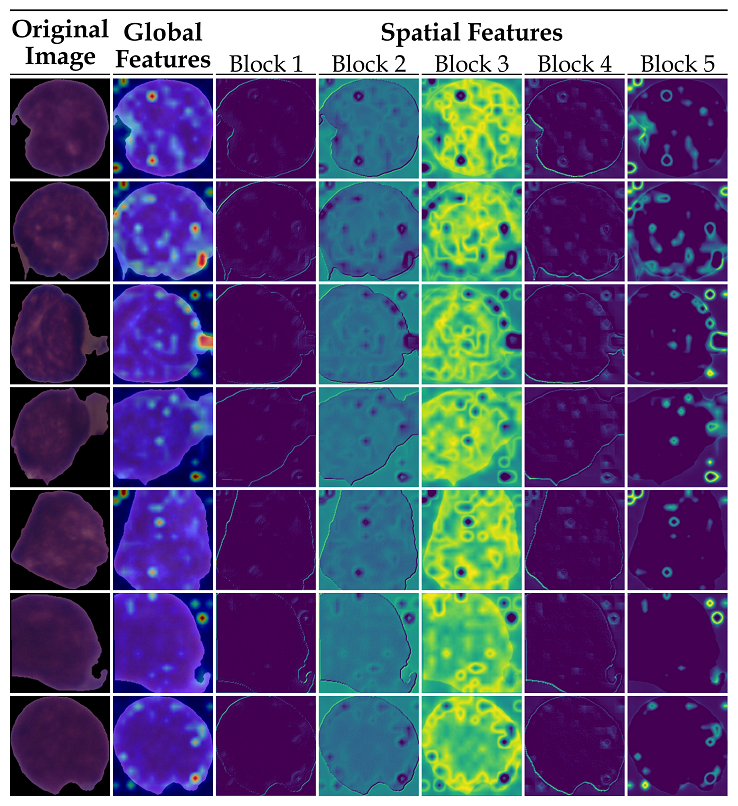}
        %\caption{default}
        \centering
        \small(b)
        %\label{fig:figure2}
    \end{minipage}
    \caption{(a) Feature maps of leukemic cells capturing global and block-wise spatial features. (b) Feature maps of normal cells capturing global and block-wise spatial features. The features of the leukemic class show denser patterns compared to the normal class. This densification pattern helps to enhance classifier performance.}
    \label{fig:fmaps}
\end{figure*}

% \begin{table*}[width=.9\textwidth,pos=h]
% \caption{Ablation Study on CNMC Dataset}
% \begin{tabular*}{\tblwidth}{@{} CLLLLL@{} }
% \toprule
%       \multicolumn{1}{c}{} & \textbf{ Methods} & \textbf{Acc$^a$} & \textbf{ Pre$^b$} & \textbf{ Rec$^c$} & \textbf{ F1$^d$} \\
%       \midrule
%       \multicolumn{1}{l}{\multirow{1}{*}{Segmentation-based Classification}} & AutoCDet{\cite{Khandekar2021}} & 0.9231 & 1.0000 & 0.8750 & 0.9333 \\
%       \midrule
%       \multicolumn{1}{l}{\multirow{1}{*}{Imbalanced Classification}} & LDet{\cite{Depto2023}} & 1.0000 & 1.0000 & 1.0000 & 1.0000 \\
    
%       \midrule
%       \multicolumn{1}{l}{\multirow{4}{*}{Deep Learning-based Classification}} & EnDet{\cite{Mohammed2023}} & 0.8462 & 1.0000 & 0.7778 & 0.8750 \\
%       \multicolumn{1}{c}{} & SDCT-AuxNet$\theta${\cite{Gehlot2020}} & 0.9231 & 0.8571 & 1.0000 & 0.9231 \\
%       \multicolumn{1}{c}{} & ALNett{\cite{Jawahar2022}} & 0.8462 & 0.8571 & 0.8571 & 0.8571 \\
%       \multicolumn{1}{c}{} & ALL-TL{\cite{Das2022}} & 0.7692 & 0.7143 & 0.8333 & 0.7692 \\
%       \midrule
%       \multicolumn{1}{c}{} & \textbf{CoTCoNet} & \textbf{1.0000} & \textbf{1.000} & \textbf{1.0000} & \textbf{1.0000} \\
%       \bottomrule
%       \multicolumn{6}{l}{\scriptsize $^a$ Accuracy, $^b$ Precision, $^c$ Recall, $^d$ F1-Score}
% \end{tabular*}
% \label{table:ablation}
% \end{table*}

\section{Discussion}
\label{discussion}
In this study, we propose an advanced framework known as CoTCoNet for detecting acute lymphoblastic leukemia. CoTCoNet leverages the power of transformers and convolutional networks to extract both global and spatial features from blood cell images. We evaluate CoTCoNet on four diverse datasets and demonstrate its superior performance and robustness over existing methods. Additionally, we employ a Local Interpretable Model-agnostic Explanations (LIME), an Explainable Artificial Intelligence (XAI) method that assigns weights to the most influential regions and highlights them. The prominently highlighted regions are the primary contributors to the accurate classification of leukocytes.

\begin{figure*}[!ht]
  \centering
  \includegraphics[width=0.95\textwidth]{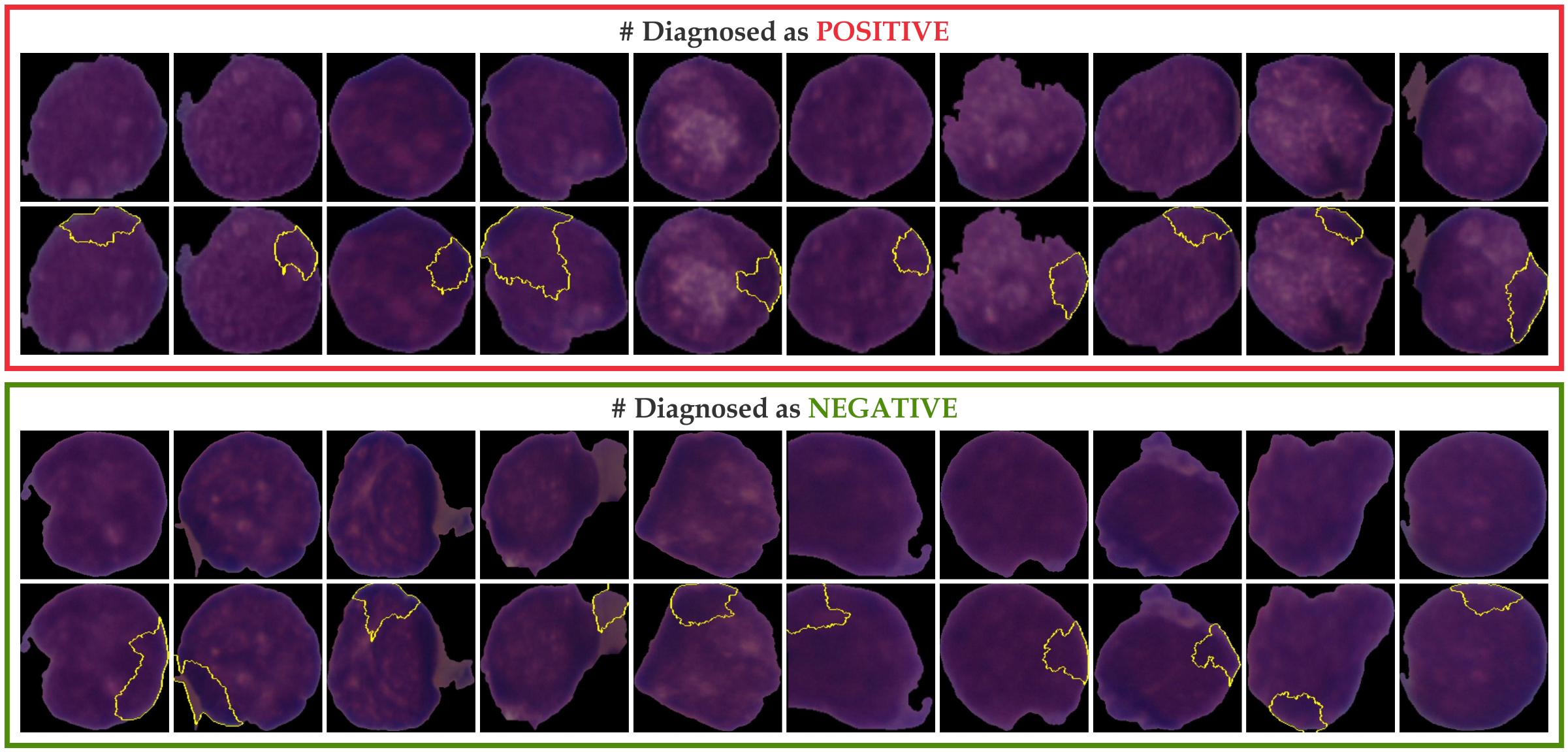}
  \caption{Visualization of 20 microscopic blood smear images from C-NMC 2019 Dataset and highlighted features using LIME. The images in the first row are diagnosed as positive (cancerous cell) by our CoTCoNet framework, while the images in the third row are diagnosed as negative (normal cell).}
  \label{fig:lime}
\end{figure*}

From a feature learning viewpoint, the transformer-based GMod module, described in \autoref{sec:gmod}, specializes in learning long-range dependencies and contextual information within the cells. Meanwhile, the CNN-based SMod module, detailed in \autoref{sec:smod}, focuses on preserving local details and the structural integrity of the cells. These coupled modules enable our framework to extract more comprehensive and discriminative features for leukemia detection. Moreover, our framework employs a novel graph-based feature reconstruction module known as GraFR, as outlined in \autoref{sec:grafr}. GraFR is a key component that identifies the hidden features and reconstructs them by computing their similarity scores with other data points in a graph structure.

\begin{figure}[!b]
  \centering
  \includegraphics[width=\columnwidth]{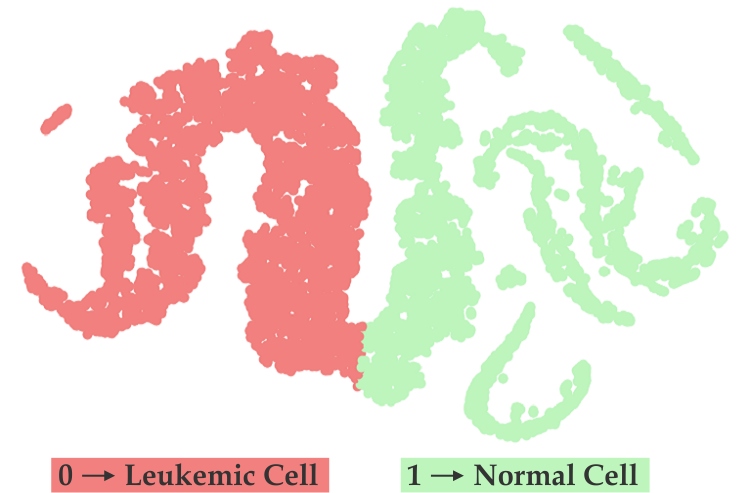}
  \caption{Class Discrimination at last layer of Hierarchical Deep Learning Classifier. The final layer allows the linear separation of the two indistinguishable classes.}
  \label{fig:classifier}
\end{figure}

The features of CoTCoNet can be visualized in~\autoref{fig:fmaps}, showing denser patterns for the leukemic class compared to the normal class. This observation implies that different densification induced by the GMod, SMod, and GraFR modules is an essential characteristic of class discrimination, where the Hierarchical Deep Learning Classifier (HDLC) is capable of effectively differentiating between the two classes based on feature density, even when the visual appearance of images from both classes is similar. \autoref{fig:classifier} illustrates class discrimination in the final layers, allowing a linear separation of the two classes. This reinforces the clear insight that the dissimilarity in density arises due to the coupled features of the CoTCoNet framework.

Classifying the leukemic cell requires robustness to subject-level variability. To assess the robustness of our framework, we conducted a comprehensive evaluation using four publicly available and diverse datasets, encompassing both single-cell and multi-cell data (refer~\autoref{sec:datasetsummary}). The experimental setup involves comparing our proposed framework with state-of-the-art methods (refer~\autoref{sec:comparison-methods}) and using appropriate evaluation metrics (refer~\autoref{sec:evaluation-metrics}) to assess performance. CoTCoNet demonstrates superior performance across diverse datasets, underscoring the profound significance of our proposed framework. To gain deeper insights into the superior performance of CoTCoNet, we conducted ablation studies (refer~\autoref{sec:ablation-study}) showing the efficacy of each module in effectively distinguishing sparse cell features.

For the visual explanation of the decision reasoning of our CoTCoNet framework, we adopt Local Interpretable Model-agnostic Explanations (LIME)~\cite{Ribeiro2016} to gain insights into the specific regions that significantly contribute to the classification process. \autoref{fig:lime} illustrates 20 microscopic blood smear images sourced from the C-NMC 2019 dataset; each image is accompanied by highlighted pertinent features generated by the LIME technique. The first ten images in the first row have been diagnosed as positive for leukemia by our CoTCoNet framework, while the images in the third row were diagnosed as negative. However, if no highly suspicious blood cells are detected, the model classifies the image as negative. This visualization underscores the interpretability of our CoTCoNet framework, providing valuable insights into its decision-making process.

\section{Conclusion}
\label{sec:conclusion}

In this study, we propose a novel framework called \textbf{Co}upled \textbf{T}ransformer \textbf{Co}nvolutional \textbf{Net}work (CoTCoNet), which hierarchically integrates a deep convolution network with a powerful transformer for the classification of leukemia within microscopic blood smear slides. CoTCoNet effectively utilizes the capabilities of transformers and deep convolution networks to proficiently extract both global and spatial features from blood cell images. Furthermore, we incorporate graph-based feature reconstruction to elevate the feature knowledge, leveraging the advantage of complex and hard-to-see biological features of the leukocytes. The aforementioned technical framework comprises several advantages: (a) the proposed method acquires features from indistinguishable cell images across different classes, significantly enhancing model flexibility in clinical practice. Moreover, the preservation of the structural integrity of cells in blood smear images yields comprehensive and discriminative features; (b) the proposed framework utilizes a Population-based Meta-Heuristic Algorithm to identify optimal features, effectively reducing optimization errors; (c) the proposed method is capable of generating interpretable attention maps to reinforce the prediction results, thereby facilitating hematopathologists in their decision-making process when utilizing AI algorithms; (d) the proposed framework demonstrates superior classification performance across four publicly available and diverse datasets.

We acknowledge some limitations in our research, firstly, the proposed method considers the final stages of leukocytes, overlooking the early stages of cellular development. Secondly, our approach depends on labeled data, necessitating considerable input from experienced experts. In our future research endeavors, we aim to overcome these limitations by developing a framework capable of identifying developmental cell states and distinguishing them based on cytochemical features. We aim to leverage prior information by integrating blood smear histology with developmental cell states to harness semi-supervised learning techniques for the automated classification of leukemia and its subtypes.

% \bibliographystyle{cas-model2-names}
\bibliographystyle{elsarticle-num}

\bibliography{cas-refs}

\end{document}